\documentclass[10pt,twocolumn,letterpaper]{article}

\usepackage{cvpr}
\usepackage{times}
\usepackage{epsfig}
\usepackage{graphicx}
\usepackage{amsmath}
\usepackage{amssymb}
\usepackage{mathtools}
\usepackage{mathabx}
\usepackage[dvipsnames]{xcolor}
\usepackage{float}
\usepackage[export]{adjustbox}
\usepackage{caption, subcaption}
\usepackage[percent]{overpic}
\usepackage{booktabs}

\usepackage[breaklinks=true,bookmarks=false]{hyperref}

\cvprfinalcopy 

\def\cvprPaperID{2923} 

\begin{document}

\title{PointRGCN: Graph Convolution Networks for 3D Vehicles Detection Refinement}

\author{Jesus Zarzar* Silvio Giancola* Bernard Ghanem\\
	King Abdullah University of Science and Technology (KAUST), Saudi Arabia \\
	{\tt\small \{jesusalejandro.zarzartorano,silvio.giancola,bernard.ghanem\}@kaust.edu.sa}
}

\maketitle

\newcommand{\mysection}[1]{\vspace{3pt}\noindent\textbf{#1}}
\newcommand{\TODO}[1]{\textcolor{red}{\textbf{\textit{[TODO] #1}}}}
\newcommand{\BG}[1]{\textcolor{red}{\textbf{\textit{[BG] #1}}}}
\newcommand{\JZ}[1]{\textcolor{red}{\textbf{\textit{[JZ] #1}}}}
\newcommand{\SG}[1]{\textcolor{red}{\textbf{\textit{[SG] #1}}}}
\newcommand{\sota}{state-of-the-art\xspace}
\newcommand{\Sota}{State-of-the-art\xspace}
\newcommand{\Table}[1]{Table~\ref{tab:#1}}
\newcommand{\Figure}[1]{Figure~\ref{fig:#1}}
\newcommand{\Equation}[1]{Equation~\eqref{eq:#1}}
\newcommand{\Section}[1]{Section~\ref{sec:#1}}
\newcommand{\job}[1]{\textcolor{red}{\textbf{\textit{[job-id] #1}}}}
\newcommand{\MRGCN}{MRGCN~\cite{li2019deepgcns_journal}\xspace}
\newcommand{\MR}{MRGCN~\cite{li2019deepgcns_journal}\xspace}
\newcommand{\EdgeConv}{EdgeConv~\cite{wang2019dynamic}\xspace}

\newcommand{\MVD}{MV3D~\cite{chen2017multi}\xspace}
\newcommand{\AVODFPN}{AVOD-FPN~\cite{ku2018joint}\xspace}
\newcommand{\AVOD}{AVOD~\cite{ku2018joint}\xspace}
\newcommand{\UberMMF}{UberATG-MMF~\cite{liang2019multi}\xspace}
\newcommand{\FPointNet}{F-PointNet~\cite{qi2018frustum}\xspace}
\newcommand{\VoxelNet}{VoxelNet~\cite{zhou2018voxelnet}\xspace}
\newcommand{\PIXOR}{PIXOR~\cite{yang2018pixor}\xspace}
\newcommand{\SECOND}{SECOND~\cite{yan2018second}\xspace}
\newcommand{\PointPillars}{PointPillars~\cite{lang2019pointpillars}\xspace}
\newcommand{\PointRCNN}{PointRCNN~\cite{shi2019pointrcnn}\xspace}
\newcommand{\FastPointRCNN}{Fast Point R-CNN~\cite{Chen2019FastPointRCNN}\xspace}
\newcommand{\STD}{STD~\cite{Yang2019std}\xspace}
\newcommand{\VoxelFPN}{Voxel-FPN~\cite{wang2019voxel}\xspace}
\newcommand{\PartA}{Part-$A^2$~\cite{shi2019part}\xspace}
\newcommand{\Patch}{Patch~\cite{lehner2019patch}\xspace}

\newcommand{\KITTI}{KITTI~\cite{Geiger2012KITTI}\xspace}
\newcommand{\KITTItrain}{KITTI \emph{train}\xspace}
\newcommand{\KITTIval}{KITTI \emph{validation}\xspace}
\newcommand{\KITTItest}{KITTI \emph{test}\xspace}

\newcommand{\A}[1]{\textcolor{Orange}{\textbf{#1}}}
\newcommand{\B}[1]{\textcolor{Green}{\textbf{#1}}}
\newcommand{\C}[1]{\textcolor{RoyalBlue}{\textbf{#1}}}
\newcommand{\D}{\bfseries}

\newcommand{\Bc}{\ensuremath{\mathcal{R}}\xspace}
\newcommand{\Ec}{\ensuremath{\mathcal{E}}\xspace}
\newcommand{\Fc}{\ensuremath{\mathcal{F}}\xspace}
\newcommand{\Gc}{\ensuremath{\mathcal{G}}\xspace}
\newcommand{\Hc}{\ensuremath{\mathcal{H}}\xspace}
\newcommand{\Nc}{\ensuremath{\mathcal{N}}\xspace}
\newcommand{\Pc}{\ensuremath{\mathcal{P}}\xspace}
\newcommand{\Rc}{\ensuremath{\mathcal{R}}\xspace}
\newcommand{\Vc}{\ensuremath{\mathcal{V}}\xspace}
\newcommand{\Wc}{\ensuremath{\mathcal{W}}\xspace}
\DeclarePairedDelimiter\ceil{\lceil}{\rceil}
\DeclarePairedDelimiter\floor{\lfloor}{\rfloor}


\begin{abstract}
    In autonomous driving pipelines, perception modules provide a visual understanding of the surrounding road scene. Among the perception tasks, vehicle detection is of paramount importance for a safe driving as it identifies the position of other agents sharing the road. In our work, we propose PointRGCN: a graph-based 3D object detection pipeline based on graph convolutional networks (GCNs) which operates exclusively on 3D LiDAR point clouds. To perform more accurate 3D object detection, we leverage a graph representation that performs proposal feature and context aggregation. We integrate residual GCNs in a two-stage 3D object detection pipeline, where 3D object proposals are refined using a novel graph representation. In particular, R-GCN is a residual GCN that classifies and regresses 3D proposals, and C-GCN is a contextual GCN that further refines proposals by sharing contextual information between multiple proposals. We integrate our refinement modules into a novel 3D detection pipeline, PointRGCN, 
    and achieve state-of-the-art performance on the easy difficulty for the bird eye view detection task.
\end{abstract}



\section{Introduction}


The task of autonomous driving has received much deserved attention in recent years.
Current progress in computer vision helps in providing reliable scene understanding from visual and geometrical data.
In particular, autonomous agents must detect vehicles, pedestrians, cyclists, and other objects on the road to ensure a safe navigation.
Pipelines for autonomous driving leverage 3D computer vision for tasks such as
object detection~\cite{chen2017multi,ku2018joint}, 
tracking~\cite{Giancola_2019_CVPR,Hu_2019_ICCV} and 
trajectory prediction~\cite{Chandra_2019_CVPR,Chang_2019_CVPR}.
%
Current state-of-the-art methods in 3D vehicle detection prefer LiDAR point clouds over images.
Performances of monocular~\cite{chen2016monocular,mousavian20173d,wang2019pseudo} and stereo-based~\cite{chen20173d,xu2018multi,Li_2019_CVPR} 3D object detection techniques are not yet comparable with LiDAR-based techniques~\cite{Yang2019std,liang2019multi}.
Recent image-based methods for 3D vehicle detection even relies on LiDAR supervision~\cite{wang2019pseudo} to generate surrogate point clouds from images.
Still, defining which point cloud representation fits the most with deep learning is not straightforward and remains an open problem~\cite{qi2017pointnet,riegler2017octnet,li2019deepgcns_journal}.
Recent advances in graph convolution networks~\cite{li2019deepgcns_journal} suggest that graph representations could provide better features for point cloud processing. 
Such a representation already outperforms the state-of-the-art in many other computer vision tasks~\cite{Wang_2019_CVPR,Nguyen_2019_ICCV,Shu_2019_ICCV,wang2019deep}. 
Thus, we investigate the use of a graph representation for LiDAR point cloud in the task of 3D vehicle detection.

\begin{figure}[t]
    \begin{subfigure}{\linewidth}
        \centering
        \begin{overpic}[height=3.6cm,cfbox=Green 3pt 0pt,trim={8cm 4cm 8cm 3cm},clip]{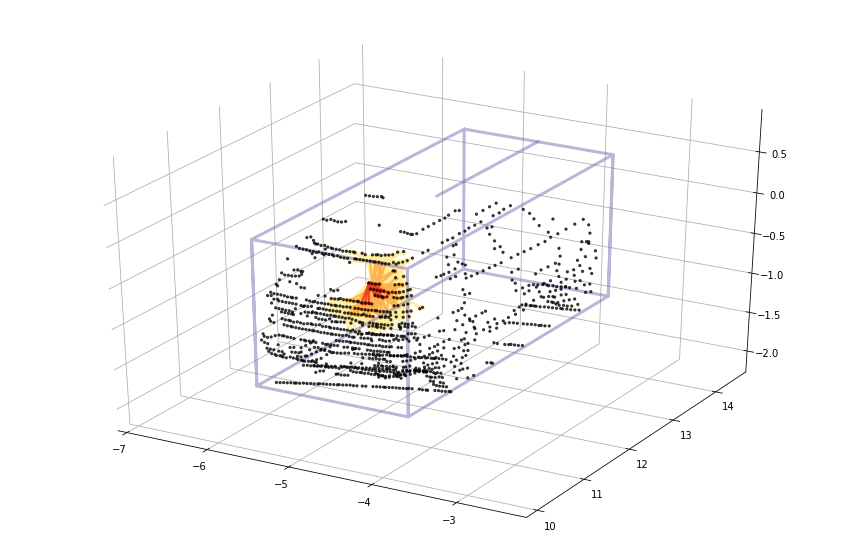}
        \put (5,78) 
        {\Large\color{Green}\textbf{R-GCN}}
        \end{overpic}
        \begin{overpic}[height=3.6cm,cfbox=RoyalBlue 3pt 0pt,trim={9cm 6cm 11cm 4cm},clip]{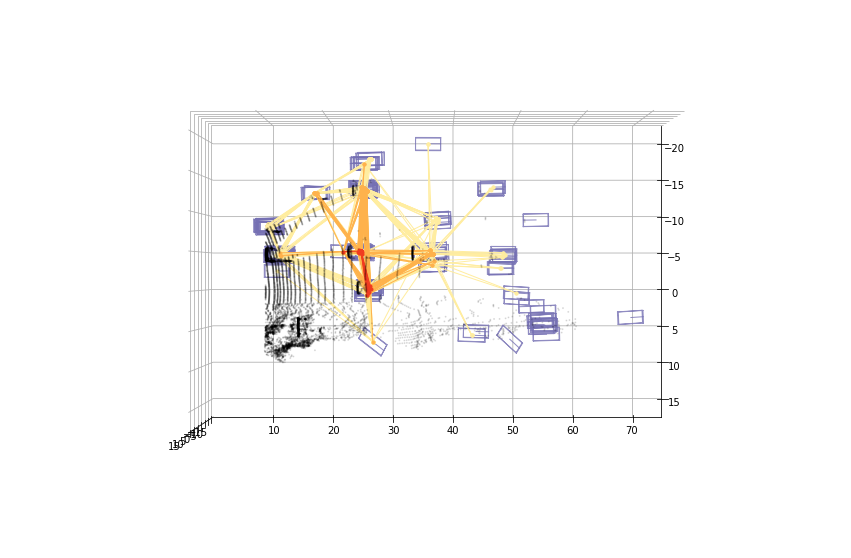}
        \put (5,84) 
        {\Large\color{RoyalBlue}\textbf{C-GCN}}
        \end{overpic}
    \end{subfigure}%
    \caption{
    \textbf{Proposed GCN modules.}
    Insights on the \textbf{\color{Orange}Receptive field} for both our proposed modules. 
    The \emph{per-proposal} \textbf{\color{Green}R-GCN} gathers meaningful features information between points on each proposal.
    The \emph{per-frame} \textbf{\color{RoyalBlue}C-GCN} gathers contextual information between proposals on each frame.
    Note that this non-contractual representation does not consider the dynamic graph on features for neighbors.
    }
    \label{fig:teaser}
\end{figure}

In our work, we propose two modules exploiting convolution operations on point cloud graph representations.
We tackle vehicle detection in a common two-stage fashion by generating a set of high-recall proposals from a LIDAR scene point clouds which we successively fine-tune to obtain refined detections.
In particular, we introduce PointRGCN: a novel pipeline consisting of two refinement modules.
R-GCN (\Figure{teaser}~(top)) provides a \emph{per-proposal} feature aggregation based on a GCN that utilizes all points contained a proposal. 
C-GCN (\Figure{teaser}~(bottom)) gathers \emph{per-frame} information from all proposals, looking for a contextual consistency in the road scene.
We will release our code after the review process.

\mysection{Contributions.} We summarize our contributions as follows.
\textbf{(i)}
We are the first, to the best of our knowledge, to leverage graph representation and GCNs for the task of 3D vehicle detection.
\textbf{(ii)} 
We propose R-GCN and C-GCN, two GCN-based modules for \emph{per-proposal} and \emph{per-frame} feature aggregation, designed to gather all features within and between proposals.
\textbf{(iii)} 
We show competitive performances for our novel pipeline PointRGCN in the challenging \KITTI 3D vehicle detection task and exhibit a $+2\%$ AP$_{BEV}$ boost on the easy subset.

\section{Related Work}

Our work relates to methods delving with 3D point cloud representations, graph convolutional networks, and 3D object detection architectures.
We present previous works on these topics, and highlight the novelty of our pipeline with respect to each of these works.

\mysection{3D Point Cloud Representation.}
Finding an efficient representation for sparse LiDAR point clouds is paramount for 3D computer vision tasks. 
Current works leverage projection, voxelization or per-point features.
Image projections allow for efficient convolution operations on the image space, and are commonplace for 3D object classification and retrieval tasks~\cite{su2015multi,qi2016volumetric,tatarchenko2018tangent}.
Projecting provides coarse but structured inputs for further deep convolution network.
Voxelization is a volumetric representation that discretizes 3D space into a structured grid.
Voxelizing 3D space allows for 3D CNN processing~\cite{wu20153d,maturana2015voxnet,riegler2017octnet}, but are not memory efficient due to the sparsity of point clouds.
PointNet~\cite{qi2017pointnet} and PointNet++~\cite{qi2017pointnet++} solve the sparsity problem by learning a deep feature representation for each point, with all points contributing to a more complete representation.
However, the context shared between points is limited to point relations in 3D space~\cite{qi2017pointnet++}, and shapes still lack structure.
In order to overcome these issues, we use GCNs as the backbone for both refinement modules in our network.

\mysection{Graph Convolutional Network.}
To cope with current limitations of point cloud representation methods, Wang~\etal~\cite{wang2019dynamic} propose a graph representation for point clouds and show improved performances in classification and segmentation tasks.
Li~\etal~\cite{li2019deepgcns_journal} show that GCNs can run as deep as 112 layers by introducing residual skip connections, similar to ResNet. 
GCNs have an increased receptive field with respect to PointNet++\cite{qi2017pointnet++} since the edge connections are dynamically assigned for each intermediate layer.
With such features, GCNs have shown success in tasks such as
segmentation~\cite{Wang_2019_CVPR},
point cloud deformation~\cite{Nguyen_2019_ICCV},
adversarial generation~\cite{Shu_2019_ICCV} and
point cloud registration~\cite{wang2019deep}.
However, To the best of our knowledge, we introduce the first 3D vehicle detection module using a graph representation on LiDAR input.

\mysection{Single-Stage 3D Object Detection.}
Single-stage detectors train a single network end-to-end to generate a small set of detection boxes with high precision.
VoxelNet~\cite{zhou2018voxelnet} directly generates detections from a Region Proposal Network using a voxelized point cloud.
SECOND~\cite{yan2018second} improves upon VoxelNet by leveraging sparse convolution operations on the voxelized input.
PointPillars~\cite{lang2019pointpillars} uses a pillars representation based on SECOND voxelisation with infinite height.
Voxel-FPN~\cite{wang2019voxel} introduces a multi-scale voxel feature pyramid network based on VoxelNet~\cite{zhou2018voxelnet}. 
Qi~\etal~\cite{Qi_2019_ICCV} recently introduced a Deep Hough Voting scheme for indoor detection, adopting PointNet++\cite{qi2017pointnet++} for the feature backbone.
%
We preferred a two-stage detector since single-stage methods tend to attain lower performances, and GCNs are computationally expensive to compute on complete point clouds.


\mysection{Two-Stage 3D Object Detection.}
On the other hand, two-stage detection methods divide the detection pipeline into two sub-modules.
One module is trained to generate a large number of high-recall proposals, and a second one to refine those proposals.
Frustum PointNet~\cite{qi2018frustum} uses 2D proposals from RGB images and localizes 3D bounding boxes using a PointNet~\cite{qi2017pointnet} representation on the 2D proposal's frustum.
Liang \cite{liang2019multi} propose a multi sensor multi view aggregation taking into account both LiDAR and RGB camera information.
Alternatively, STD~\cite{Yang2019std} defines spherical anchors to obtain higher recall in the proposal stage.
Chen~\etal~\cite{Chen2019FastPointRCNN} use a voxelization for their Region Proposal Network (RPN), and use PointNet features for proposal refinement.
Also, the concurrent work of Lehner~\etal~\cite{lehner2019patch} proposes a coarse RPN on the top view projection which allows for a denser voxelization of proposals during refinement.
%
PointRCNN~\cite{shi2019pointrcnn} creates object proposals by first performing point segmentation using a PointNet++ backbone, regressing a proposal for each foreground point, and finally refining the proposal using PointNet++. 
Part-$A^2$ Net~\cite{shi2019part} improves upon PointRCNN by adding a part aware loss for intra-proposal point locations, and performing proposal refinement by using a Voxel Feature Encoding.
%
%

In our work, we build upon \PointRCNN by using its proposal scheme, but leverage GCNs to refine proposals.
Due to the computational and memory complexity of GCNs, we limit their usage to the refinement stage.
In particular, we combine point features for each proposal with R-GCN, and aggregate proposal context information for each frame with C-CGN.

\section{Methodology}

In this section, we present our 3D object detection pipeline, PointRGCN, which takes advantage of recent advances in GCNs.
Our model takes as an input a LIDAR scan of a scene and generates tight 3D bounding boxes for cars found in the LIDAR point cloud.
Similar to common practice, we split object detection into two stages: generating high recall proposals, and refining the set of proposals to get detections with high precision.
The main focus of our network is on the second stage of object detection: refining proposals.
Therefore, we take the region proposal network from an existing method, PointRCNN~\cite{shi2019pointrcnn}, to generate proposals.
During refinement, we take advantage of geometric information contained in point clouds by leveraging a graph representation for points within proposals.
Furthermore, we aggregate the contextual information across proposals in a second refinement step.
In the following section, we present an introduction to GCNs and how they are applied in the context of object detection.
We then present the details of our pipeline PointRGCN, including the two main components of our refinement network: \textbf{R-GCN}, and \textbf{C-GCN}.

\begin{figure*}[t]
    \centering
    \begin{overpic}[height=2.13cm,cfbox=Black 2pt 0pt,
    trim={12cm 5cm 7cm 8cm},clip]{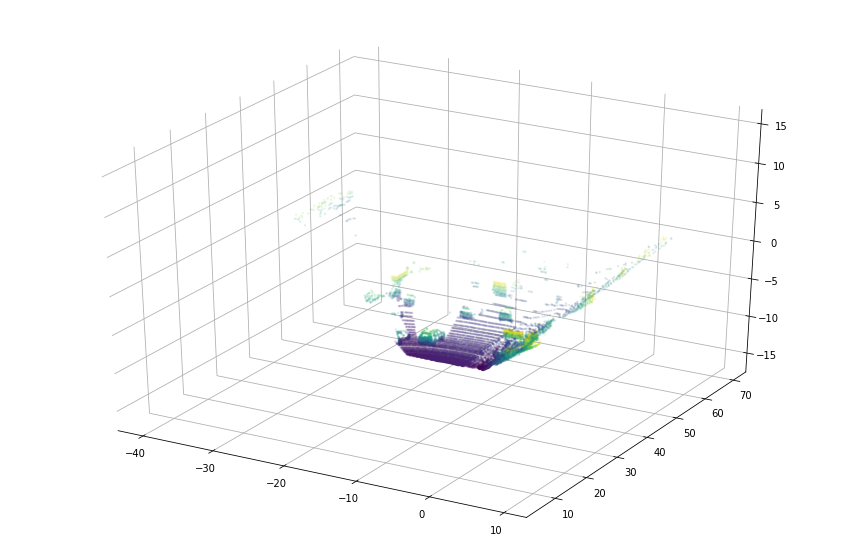}
    \put (4,5) {\large\color{Black}\textbf{Point Cloud}}
    \end{overpic}
    \begin{overpic}[height=2.13cm,cfbox=Orange 2pt 0pt, 
    trim={12cm 5cm 7cm 8cm},clip]
    {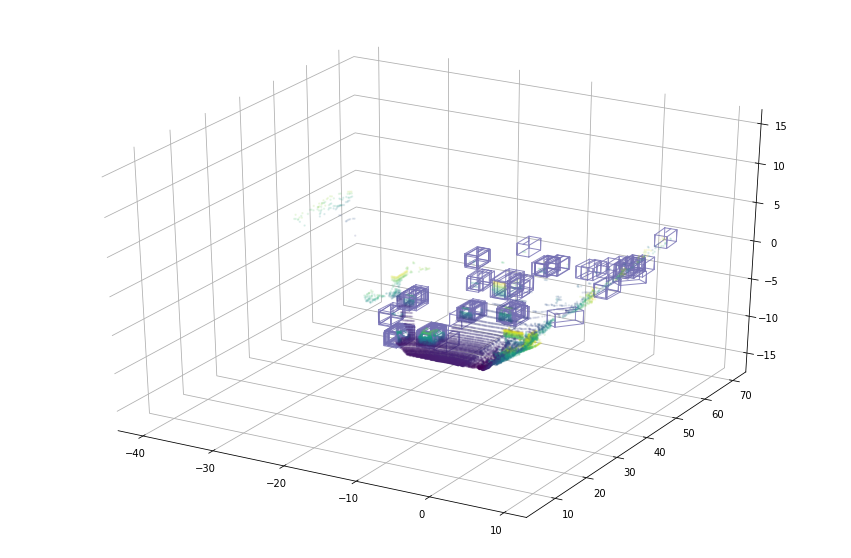}
    \put (4,5) {\large\color{Orange}\textbf{Proposals}}
    \end{overpic}
    \begin{overpic}[height=2.13cm,cfbox=Green 2pt 0pt, 
    trim={8.5cm 3cm 8cm 4cm},clip]
    {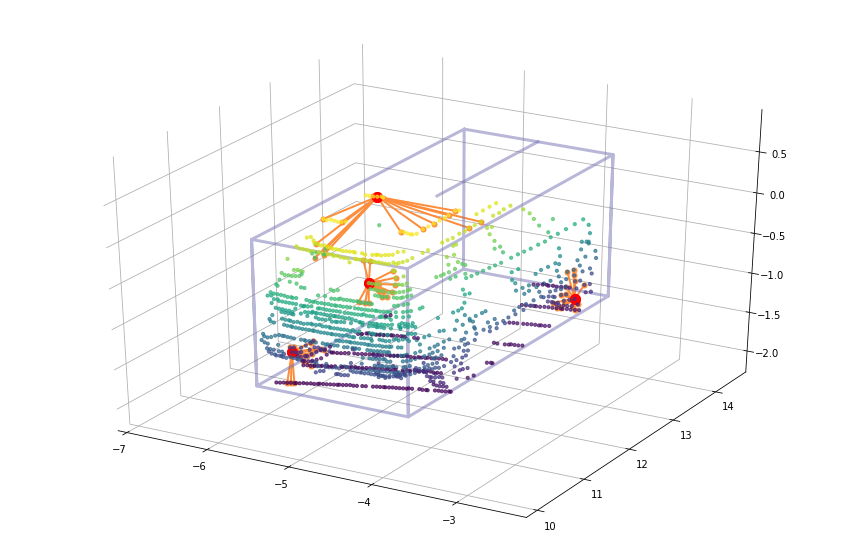}
    \put (5,6) {\large\color{Green}\textbf{R-GCN}}
    \end{overpic}
    \begin{overpic}[height=2.13cm,cfbox=RoyalBlue 2pt 0pt, 
    trim={12cm 5cm 7cm 8cm},clip]
    {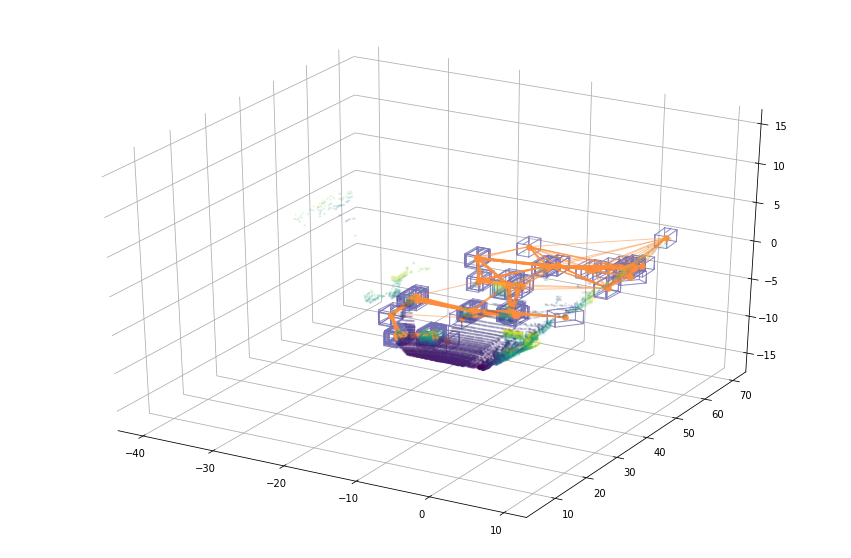}
    \put (4,5) {\large\color{RoyalBlue}\textbf{C-GCN}}
    \end{overpic}
    \begin{overpic}[height=2.13cm,cfbox=Black 2pt 0pt, 
    trim={12cm 5cm 7cm 8cm},clip]
    {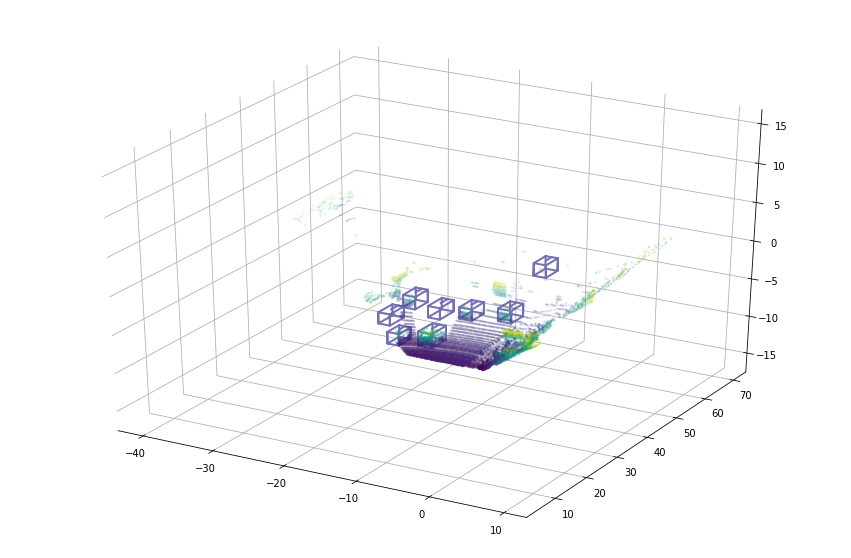}
    \put (4,5) {\large\color{Black}\textbf{Detection}}
    \end{overpic}
    \caption{
    \textbf{PointRGCN:}
    We propose a novel object detection pipeline that introduce Graph Convolution Networks (GCNs) in the refinement module.
    {\color{orange}\textbf{Proposals}}: We generate proposals by regressing a bounding box per each foreground vehicle points, similar than PointRCNN~\cite{shi2019pointrcnn}.
    {\color{ForestGreen}\textbf{R-GCN}}: We improve the per-proposal feature extraction used to classify and regress the 3D object proposals by introducing residual GCNs.
    {\color{blue}\textbf{C-GCN}}: We share the features between proposals to embed a contexual consistency between the objects to detect.
    }
    \label{fig:pipeline}
\end{figure*}

\subsection{Graph Embedding and GCNs.}

\mysection{Definitions.}
A graph \Gc is defined as a tuple consisting of a set of nodes and a set of edges, $(\Vc, \Ec)$, in which an edge $e_{i,j}\in\Ec$, between nodes $v_i\in\Vc$ and $v_j\in\Vc$ takes a binary value signifying whether node $v_i$ is related to node $v_j$ or not.
Connections between nodes are directed, \ie an edge $e_{i,j}$ may differ from its reciprocal edge $e_{j,i}$.
Each node is represented by a $c$-dimensional feature vector, \ie $v_i \in \mathbb{R}^c$.

\mysection{Graph Convolutional Networks.}
GCNs are an extension of CNNs which operate on graph representations instead of regular image grids.
Given an input graph \Gc, a graph convolution operation \Fc aggregates features from $k$ nodes in a neighborhood $\Nc^{(k)}(v)$ of a given node $v$.
A convolution operation \Fc updates the value of the given node by aggregating features amongst its neighbor nodes, as presented in \Equation{GCN}.
This aggregates features from nearby nodes, mirroring the way convolutional filters aggregate nearby pixels.
Note that unlike CNNs, GCNs do not apply different weights to different neighbors. 
%

\begin{equation}
\begin{aligned}
    \Fc (\Gc_l,\Wc_l )& = \text{Update} (\text{Aggregate} ( \Gc_l  , \Wc_l^{a} ) , \Wc_l^{u} ) \\
    \Gc_{l+1} & = \Fc ( \Gc_l , \Wc_l ) + \Gc_l
    \label{eq:GCN}
\end{aligned}
\end{equation}



\mysection{EdgeConv.}
Node features are aggregated around neighbors $\Nc^{(k)}(v_l)$ of nodes $v_l$ at layer $l$.
In particular, the feature difference $\mathbf{h}_{u_l}-\mathbf{h}_{v_l})$, for given neighbor features $u_l\in\Nc^{(k)}(v_l)$ is concatenated to the current node's features $\mathbf{h}_{v_l}$, and processed with a shared multi-layered perceptron (MLP).
This makes use of both a local geometry encoded by the feature difference as well as a global geometry obtained from the features of each node center $\mathbf{h}_{v_l}$.
Then, a max operation pools features between the neighbors and updates the current node's feature $\mathbf{h}_{v_{l+1}}$ for next layer $l+1$.
\Equation{EdgeConv} illustrates the aggregation and update steps introduced by \EdgeConv.


\begin{equation} 
\begin{aligned} 
    \mathbf{h}_{v_{l+1}} &
    =\max(\{\mathbf{h}(u_l)|u_l\in\Nc^{(k)}(v_l)\})\\
    \mathbf{h}(u_l) &
    =\text{MLP}(\text{concat}(\mathbf{h}_{v_l},\mathbf{h}_{u_l}-\mathbf{h}_{v_l}))
    \label{eq:EdgeConv}
\end{aligned} 
\end{equation}

\mysection{MRGCN.}
Rather than performing the MLP operation on all neighbors before the maxpool operation in \Equation{EdgeConv}, \MRGCN performs the maxpool first, and applies the MLP only once.
%
In particular, the difference features $\mathbf{h}_{u_l}-\mathbf{h}_{v_l})$ for all neighbor features $u_l\in\Nc^{(k)}(v_l)$ are maxpooled into an intermediate representation $\mathbf{h}_{\Nc^{(k)}\left(v_l\right)}$. 
Then, that representation is concatenated with the current node feature and processed once with an MLP.
This reduces the memory budget since a single intermediate feature vector is stored for each neighbor, rather than $k$ of them.
Also, the MLP is performed only once, allowing for faster inference.


\begin{equation} 
\begin{aligned} 
    \mathbf{h}_{v_{l+1}}&
    =\text{MLP}\left(\text{concat}\left(\mathbf{h}_{v_l},\mathbf{h}_{\Nc}\left(v_l\right)\right)\right)\\
    \mathbf{h}_{\Nc^{(k)}}\left(v_l\right)&
    =\max(\{\mathbf{h}_{u_l}-\mathbf{h}_{v_l}|u_l\in\Nc^{(k)}\left(v_l)\}\right)
    \label{eq:MRGCN}
\end{aligned}
\end{equation}

\mysection{Dynamic Graphs.}
Dynamic GCNs recompute a new graph per layer using the feature space produced by each intermediate representation~\cite{wang2019deep}.
In particular, edges $e_{i,j}$ are recomputed in each layer by taking $k$ closest nodes using a Euclidean distance in the current layer's feature space $\mathbf{h}$.
By using dynamic graph updates, the receptive field becomes dynamic.
This allows relations to be drawn between far apart nodes, given they have similarities, and thus provides additional contextual information.




\mysection{Dilated GCNs.}
Convolutional kernel dilations showed improvements in CNNs~\cite{yu2015multi}, and have recently been extended into graph convolutions as well.
In the convolution domain, dilation increases a convolutional filter's receptive field without increasing the number of model parameters.
Dilation is achieved by using a larger convolution kernel, but fixing every other kernel coefficient to $0$.
Similarly, dilation on GCNs is performed by skipping neighbors to increase the receptive field~\cite{Li2019deepgcn}.
In this manner, further neighbors can be reached without increasing the total number of neighbors aggregated in each graph convolution.
A dilation of $d$ implies that every $d$-th neighbor is used for aggregation, which effectively increases the receptive field to $d \times k$ neighbors at a given layer.
We use a linearly increasing dilation with respect to the number of layers, \ie the $l$-th layer uses a dilation of $l$.

\mysection{Residual Skip Connections.}
In a similar manner, residual skip connections have been adopted from CNNs.
We apply residual skip connections by adding the previous graph nodes features to the next layer, providing a better gradient flow and faster convergence.

\subsection{PointRGCN Vehicle Detection Pipeline}
Our pipeline is composed of $3$ main modules: a region proposal network, a graph-based proposal feature extraction network, and a graph-based proposal context aggregation network.
The input to our network is a scene point cloud, and the output is a set of refined boxes \Rc (\Figure{pipeline}).

\mysection{RPN: Proposal Generation.}
This module takes as input the scene point cloud, and generates a set of proposal bounding boxes \Bc.
Proposal boxes $b^i$ are defined by a $7$-dimensional vectors consisting of a center, dimensions, and a rotation along the vertical axis: $b^i = (x, y, z, h, w, l, \theta)$.
The main requirement for this module is to have a high recall for the set of proposals \Bc in the input scene.
In our work, we leverage proposals from~\cite{shi2019pointrcnn}.
We do not alter on the proposal generation as the average recall of their method is relatively high already.
This proposal framework first learns to segment foreground and background points, and extrapolates a proposal bounding box $b^i$ for each foreground point.
Both proposals and segmentation features are used further down the pipeline during proposal refinement.
Current GCN-based feature extraction methods are memory intensive hence not suitable for feature extraction from large point clouds such as an outdoor scene captured with LIDAR.
As such, we rely on PointNet++~\cite{qi2017pointnet++} to extract meaningful per-point features for the segmentation performed in this stage.

\mysection{R-GCN: Proposal Feature Extraction.}
The aim of this network is to take a set of proposal boxes \Bc with a high recall, and extract a set of features from each proposal.
These features are later on used to regress better bounding boxes.
To achieve this, proposal boxes are first expanded to include extra context from objects surrounding the proposal.
LIDAR points lying within each of the expanded proposal boxes are cropped, and a total of $P$ points are kept to form the set of points $\Pc^i$ for any proposal $i$.
The isolated proposal point clouds are transformed to a canonical reference frame where the proposal's center lies at the origin, and where the proposal axes are aligned with the canonical frame's axes.
For each point, we project its canonical coordinates into a larger space to match the dimensionality of the point's RPN features.
We then concatenate the projected canonical coordinates with the corresponding RPN features, and reduce the concatenated vector's dimensionality in order to obtain a per-point feature vector.
Our R-GCN module processes these per-point feature vectors using $N_r$ layers of \MRGCN. 
Each layer has a fixed number $F_r$ of filters, and residual skip connections are used between consecutive layers.
We then create a local multi-layer feature vector for each point by concatenating the output features from every graph convolution layer.
For each proposal, we learn a global point feature by projecting the local multi-layer feature into a $C_r$-dimensional space, and performing a max pool across all points.
The global point feature is then concatenated to every point in each proposal.
Finally, a max pool is performed across all points for each proposal box.
This generates an $(N_r \times F_r + C_r)$-dimensional output feature vector that captures both global and local information within proposals.


\mysection{C-GCN: Context Aggregation.}
We would like to exploit the fact that most proposals will be related by physical (\eg ground plane) and road (\eg lanes) constraints.
For this purpose, we leverage the information encapsulated in each proposal features to further refine all proposals.
We gather the set of $\vert \Bc \vert$ proposal feature outputs from our R-GCN module into a graph representation of the current frame.
In particular, each node on the graph represents a proposal with its R-GCN feature representation.
We then process the proposal graph with $N_c$ layers of \EdgeConv, each with $F_c$ filters, and residual skip connections between consecutive layers.
We compute a global feature like for R-GCN and concatenate it to each proposal's local feature vector,  producing an output of dimension $(N_c \times F_c + C_c)$ with aggregated proposal information.

\subsection{Detection Prediction}
In the last module, the feature vector generated for each proposal is used to regress a refined detection box along with a classification score for the given proposal.
Two sets of fully connected layers are used for this purpose: one for classification, and one for regression.
The purpose of classification is to predict whether a given proposal is good enough.
Regression is performed in two different manners specified below: binned regression, and residual regression as is done in~\cite{shi2019pointrcnn}.
Binned regression is performed for $f \in \{x,z,\theta\}$, while residual regression is performed for box features $f \in \{y,h,w,l\}$.

\mysection{Classification.}
Proposal classification is performed by predicting a confidence score of whether a proposal is good enough to predict a refined box or not.
It is assumed that a proposal which has a large IoU with a ground truth bounding box should lead to an even better refined detection.
Therefore, a proposal is considered positive if its IoU with the closest ground truth bounding box is larger than a set threshold, and negative if its IoU is lower than a different threshold.
Proposals in the gray area between the positive and negative thresholds are not used during training.
This classification score is used to determine which detection should be kept during inference.

\mysection{Binned Regression}
Binned regression is done by performing a classification among a set $V_f$ of different bins for each box feature $f$ using binned regression.
Alongside this bin classification, an intra-bin regression is also performed to allow for finer box prediction.
Bin centers $v^i_f$ for bin indices $i \in [0,\vert V_f \vert)$ are calculated as shown in \Equation{refine}, for a given feature $f \in \{x,z,\theta\}$.
Here, $S_f$ is the search space which is being discretized into bins, $\delta_f$ is the bin resolution, and $v_f$ is the proposal box feature.
The bin center with the highest classification score, $\hat{v}_f$, is used to generate the refined box.
Refined box features $\hat{r}_f$ are calculated as shown in \Equation{refine}, where $\widehat{breg}_f$ is the network regression output corresponding to the bin with the highest score.
The bin size $\delta_f$ is used to normalize regression outputs in order have a more stable training.




\begin{equation}
    \begin{split}
    v^i_f &= b_f + (i + 0.5) \delta_f  - S_f\\
    \hat{r}_f &= \hat{v}_f + \delta_f \widehat{breg}_f\\
    \label{eq:refine}
    \end{split}
\end{equation}

\mysection{Residual Regression.}
Residual regression can be thought of as a binned regression with a single bin except that bin classification is no longer necessary since there is a single bin.
Thus, only one regression value $\widehat{breg}_f$ is calculated by the network for each box feature $f$ that is being regressed in this manner.
Refined box features are obtained in the same manner as with binned regression, as shown in \Equation{refine}.
However, as bins are no longer being calculated, $\hat{v}_f$ and $\delta_f$ take new meanings with some subtle differences between the case when $f \in \{y\}$, and the case when $f \in \{h, w, l\}$.
When $f \in \{y\}$, $\hat{v}_f = b_f$, and $\delta_f = 1$.
This means the single bin center is located at the proposal's $y$ location, and there is no normalization of the calculated regression value.
If $f \in \{h, w, l\}$, both $\hat{v}_f$ and $\delta_f$ are taken to be the mean of the box feature among all training samples \ie $\hat{v}_f = \delta_f = \{1.53, 1.63, 3.88\}$, for $f \in \{h, w, l\}$.
Therefore, both the single bin's center and size are fixed to the mean of the corresponding box dimension.



\subsection{Losses}
Our final loss is composed of two main components, proposal classification $ \mathcal{L}_{cls} $ and proposal regression $ \mathcal{L}_{reg} $, which we minimize jointly.
The proposal regression loss can be further decomposed into a binned classification loss, and a binned regression loss.
The loss equations used for both classification and regression are shown in \Equation{Losses}.
A binary cross entropy loss is used for both proposal and bin classifications while a
smooth L1 loss is used for the regression tasks, \ie $\mathcal{F}_{cls}$ is the binary cross entropy loss, and $\mathcal{F}_{reg}$ is the smooth L1 loss.
$\mathcal{B}_{reg}$ is the subset of proposals with an IoU large enough to be used for regression.

\begin{equation}
    \begin{split}
        \mathcal{L}_{cls} = &\frac{1}{\vert\mathcal{B}\vert} \sum_b{\mathcal{F}_{cls}(\widehat{cls}^b, \widebar{cls}^b)}  \\
        \mathcal{L}_{reg} = &\frac{1}{\vert\mathcal{B}\vert} \sum_{b \in \mathcal{B}_{reg}} \sum_{f \in \{x,z,\theta\}}\mathcal{F}_{cls}(\widehat{bcls}^b_f, \widebar{bcls}^b_f)\\ 
        + &\frac{1}{\vert\mathcal{B}_{reg}\vert} \sum_{b \in \mathcal{B}_{reg}} \sum_f\mathcal{F}_{reg}(\widehat{breg}^b_f, \widebar{breg}^b_f) 
    \end{split}
    \label{eq:Losses}
\end{equation}

\mysection{Regression Targets}
We compute the targets for regression as follows.
The ground truth bin for a given proposal $b$ and box feature $f$ can be computed as shown in Equation~\eqref{eq:gtBin}, where $\bar{b}_f$ is the ground truth bounding box with the highest IoU for proposal $b$.

\begin{equation}
    \begin{split}
    \widebar{bin}^b_f &= \floor*{\frac{\bar{b}_f - b_f + S_f}{\delta_f}} \\
    \end{split}
    \label{eq:gtBin}
\end{equation}

Binned regression classification targets are the one-hot encoded vector of ground truth bins $\widebar{bin}^b_f$ amongst all $\| \mathcal{V_f} \|$ possible bins, \ie $\widebar{bcls}^b_f = \text{onehot}(\widebar{bin}^b_f)$.
Once the ground truth bin has been computed for a given feature, its regression target can be computed as shown in \Equation{regressionTargets}, where $\bar{v}_f$ is the ground truth bin center.
In the cases where there is a single bin, the ground truth bin center is equal to the single bin center \ie $\bar{v}_f = \hat{v}_f$.
It is worth noting that regression is only performed for the single ground truth bin for each proposal.

\begin{equation}
    \begin{split}
    \widebar{breg}^b_f &= \frac{\bar{b}_f - \bar{v}_f}{\delta_f} \\
    \end{split}
    \label{eq:regressionTargets}
\end{equation}



\section{Experiments}
\label{sec:Experiments}
In this section, we detail the experiments performed to validate our pipeline.
We first present the experimental setup, including the dataset, network, and training details.
Afterwards, we show our main results on KITTI testing set and compare our pipeline with other state-of-the-art in both 3D object detection and BEV object detection.
Finally, we present ablation studies performed on different model components to validate our pipeline design choices.

\begin{table*}[t]
	\centering
	\caption{
		\textbf{Main Results on KITTI \emph{testing} set.}
		We report metrics published in papers.
		The first $4$ methods leverage LiDAR and RGB information,
		while the next $7$ LiDAR only.
		For each columns, we highlight the \A{first}, \B{second} and \C{third} best published method using LiDAR only.
		Our method performs best on the easy difficulty for AP$_{BEV}$.
	}
	\label{tab:MainResults}
	\begin{tabular}{l|c||c|c|c||c|c|c||c}
		&          & \multicolumn{3}{c||}{3D @ 0.7 IoU}& \multicolumn{3}{c||}{BEV @ 0.7 IoU} & Time        \\ \hline
		Method    & Modality &   Easy  &   Mode. &   Hard  &   Easy  &   Mode. &   Hard  & (ms)   \\ \hline\hline
		\MVD      &      L+I &   66.77 &   52.73 &   51.31 &   85.82 &   77.00 &   68.94 &  240  \\ \hline
		\AVOD     &      L+I &   73.59 &   65.78 &   58.38 &   86.80 &   85.44 &   77.73 &  100  \\ \hline
		\AVODFPN  &      L+I &   81.94 &   71.88 &   66.38 &   88.53 &   83.79 &   77.90 &  100   \\ \hline
		\FPointNet&      L+I &   81.20 &   70.39 &   62.19 &   88.70 &   84.00 &   75.33 &  170   \\ \hline
		\UberMMF  &      L+I &   86.81 &   76.75 &   68.41 &   89.49 &   87.47 &   79.10 &   80   \\ \hline\hline
		\VoxelNet &        L &   77.49 &   65.11 &   57.73 &   89.35 &   79.26 &   77.39 &  220   \\ \hline
		\PIXOR    &        L &    -    &    -    &    -    &   84.44 &   80.04 &   74.31 &  100   \\ \hline
		\SECOND   &        L &   83.13 &   73.66 &   66.20 &   88.07 &   79.37 &   77.95 &   50   \\ \hline
		\PointPillars &    L &   79.05 &   74.99 &   68.30 &   88.35 &\C{86.10}&   79.83 &   16  \\ \hline
		\PointRCNN &       L &\C{85.94}&\B{75.76}&   68.32 &   89.47 &   85.68 &   79.10 &  100   \\ \hline
		\FastPointRCNN &   L &   84.28 &\C{75.73}&   67.39 &   88.03 &\C{86.10}&   78.17 &   65   \\ \hline
		\STD      &        L &\A{86.61}&\A{77.63}&\A{76.06}&\C{89.66}&\A{87.76}&\A{86.89}&   80   \\ \hline\hline
		R-GCN only (ours)& L &   83.42 &   75.26 &\C{68.73}&\A{91.91}&   86.05 &\B{81.05}&  239   \\ \hline  
		PointRGCN (ours) & L &\B{85.97}&\C{75.73}&\B{70.60}&\B{91.63}&\B{87.49}&\C{80.73}&  262   \\ \hline \hline %
	\end{tabular}
\end{table*}

\begin{figure*}[!htbp]
    \centering
    \frame{\begin{overpic}[width=0.33\linewidth,trim={4cm 23cm 4cm 10cm},clip]{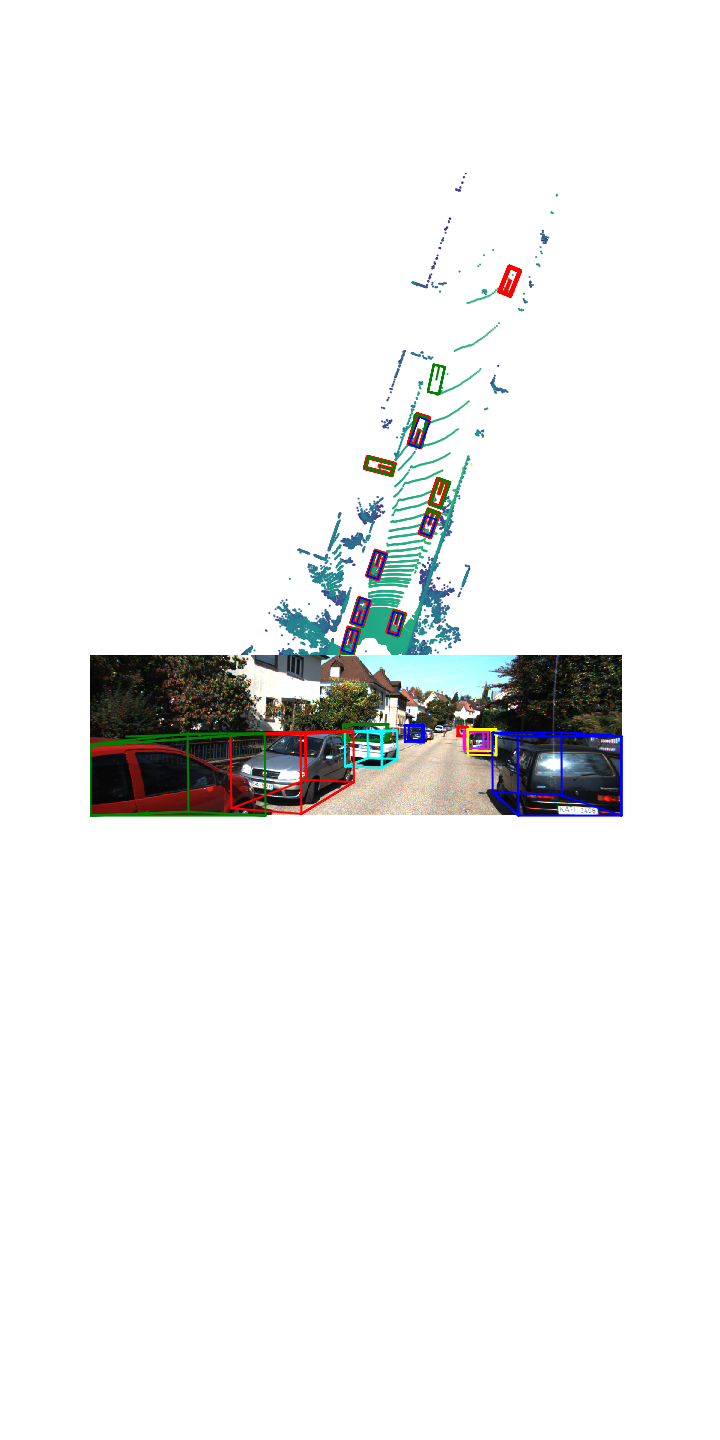}
        \put (5,88) {\LARGE\color{RoyalBlue}\textbf{(a)}}
    \end{overpic}}
    \frame{\begin{overpic}[width=0.33\linewidth,trim={4cm 23cm 4cm 10cm},clip]{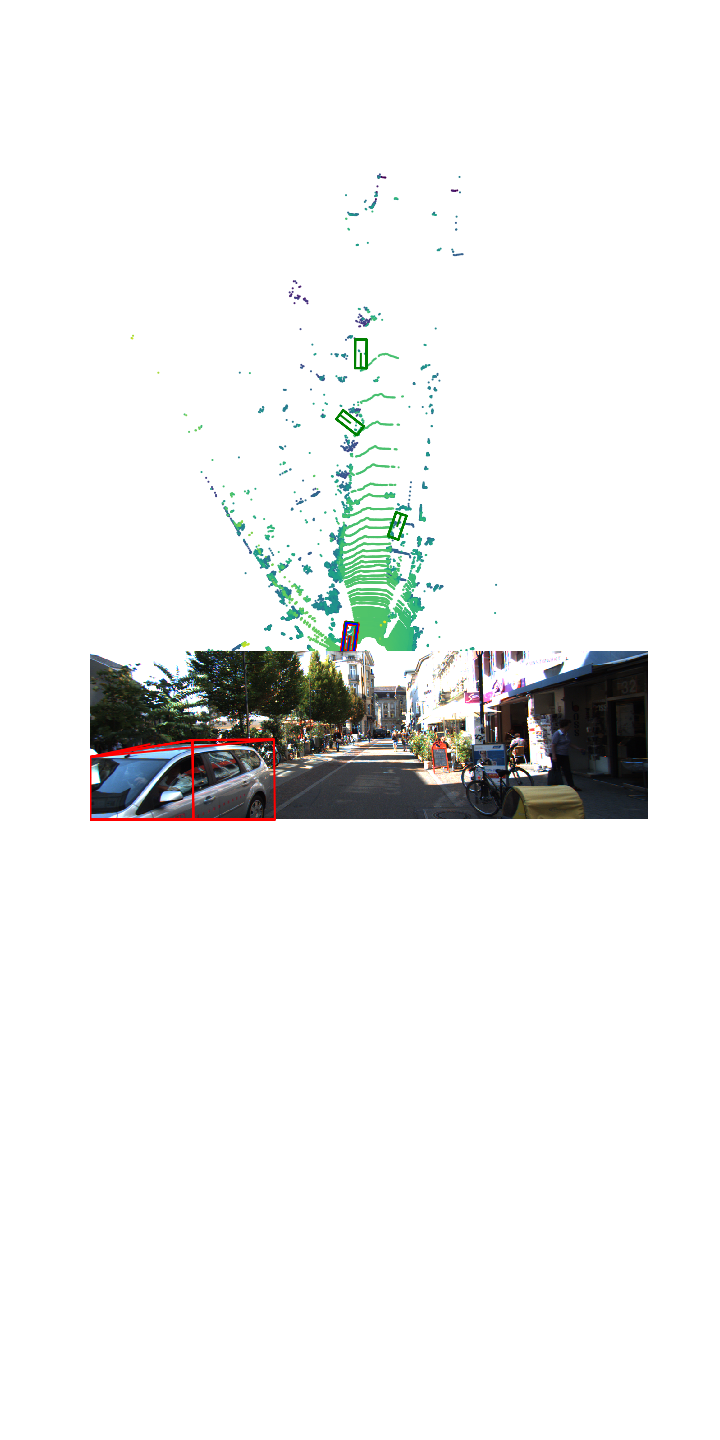}
        \put (5,88) {\LARGE\color{RoyalBlue}\textbf{(b)}}
    \end{overpic}}
    \frame{\begin{overpic}[width=0.33\linewidth,trim={4cm 23cm 4cm 10cm},clip]{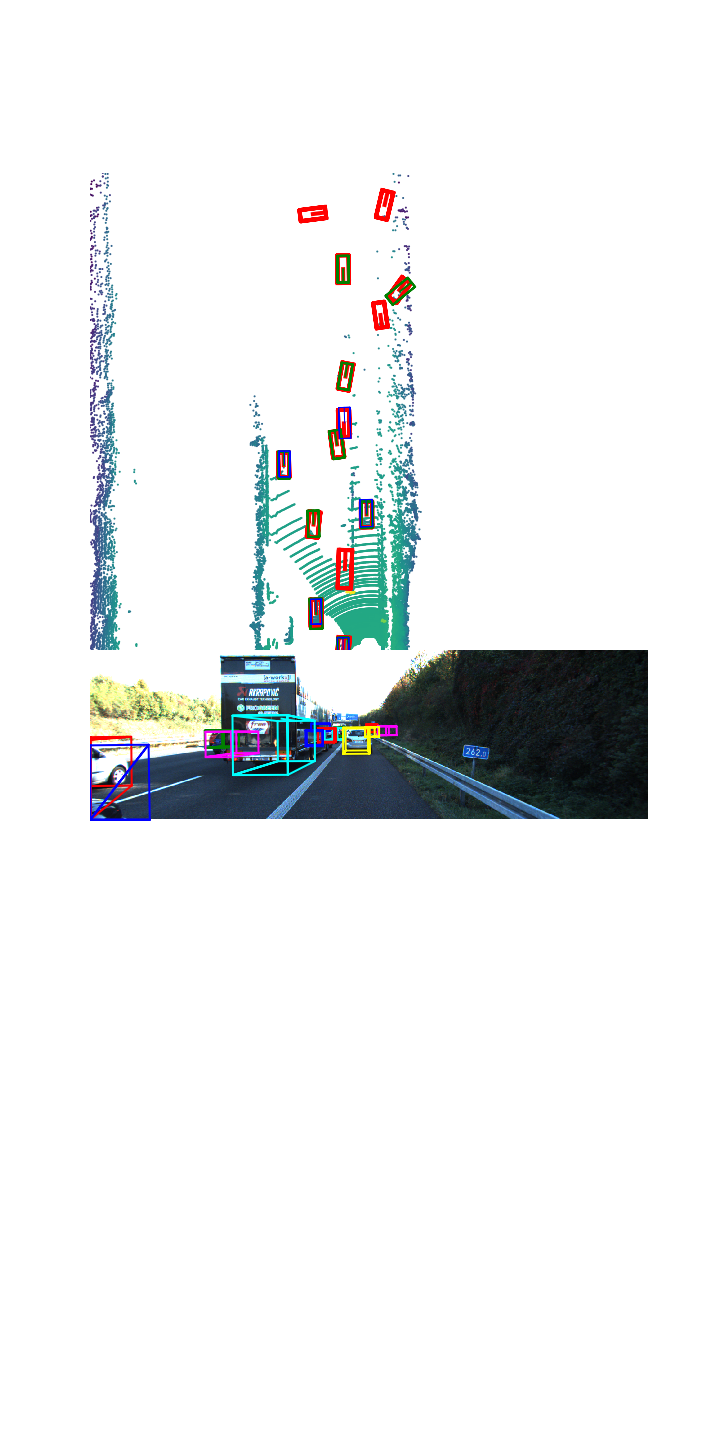}
        \put (5,88) {\LARGE\color{RoyalBlue}\textbf{(c)}}
    \end{overpic}}
    \caption{
    \textbf{Qualitative Results:}
    \textbf{Top:}
    \textbf{\color{RoyalBlue}Ground truth},
    \textbf{\color{Green}PointRCNN}~\cite{shi2019pointrcnn} and
    \textbf{\color{Red}PointRGCN [ours]} detection results on BEV projections.
    \textbf{Bottom:}
    We project the detections from \textbf{\color{Red}PointRGCN [ours]} onto the RGB image plane for visualization. \textbf{(a)} showcases where our pipeline is able to detect unlabeled vehicles from the dataset, \textbf{(b)} showcases where our pipeline is able to avoid false positives compared with PointRCNN, and \textbf{(c)} shows failure cases for our pipeline.
    }
    \label{fig:Qualitative_Detection}
\end{figure*}

\subsection{Experimental setup}

\mysection{Dataset.}
We perform our experiments on the KITTI dataset~\cite{Geiger2012KITTI} for 3D object detection, where detections are divided into $3$ difficulty levels: easy, moderate, and hard, based on occlusion and distance with respect to the vehicle's camera.
There are three major labeled object classes: cars, pedestrians, and cyclists.
We report results only on the car category since it has the highest number of training samples.
Note that the same pipeline is applicable to other categories without further modifications.
The publicly available training set is divided into training and validation subset with $3712$ and $3769$ samples each as common practice, and all results reported are taken from the validation subset except for the final testing results.

\mysection{R-GCN.}
We refine proposals by considering their canonical representation, centered on the bounding box reference system.
Proposal bounding boxes are extended by $1$ meter in every direction before being cropped, and $512$ points are randomly sampled for each proposal to allow for batch training.
We use $N_r=5$ residual \MRGCN layers with $F_r=64$ filters each and a linearly increasing dilation.
We look for $k=16$ neighbors per node and use $C_r=1024$ for the global feature size.
We use an aggregation of both canonical point coordinates and RPN features as input for each node.
We report ablation studies on the type of convolution layer (\ie \EdgeConv vs \MRGCN), network depth, the use of residual skip connections, the use of dilation and the importance of input RPN features in \Table{AblationRGCN}.

\mysection{C-GCN.}
We gather contextual information across proposals in order to detect more consistent and coherent bounding boxes.
For each proposal, we maxpool the R-GCN point features to define each node features.
We define the edges of the graph by dynamically looking for the closest $k=16$ neighbors in the feature space.
We use $N_c = 3$ residual \EdgeConv layers with $F_c=64$ filters without any dilation.
We compute a global feature of size $C_c=1024$.
In our complete PointRGCN pipeline, we use the features from R-GCN, but also experiment with the the RCNN module introduced in \PointRCNN, to highlight the contribution of our sole additional C-GCN.
We experiment with the type of layer (\EdgeConv vs \MRGCN), the depth of the network, the residual skip connections, dilation and additional RPN features aggregation. 
We report these ablation studies in \Table{AblationCGCN}.

\mysection{Bin Parameters.}
For location coordinates $f \in {x,z}$, we set the search space $S_f = 1.5m$, and the bin resolution $\delta_f = 0.5m$.
This leads to a total of $6$ bins for $f \in {x,z}$.
In the case of $f = \theta$, we use a search space $S_f = 22.5^\degree$, and bin resolution $\delta_f = 5^\degree $, for a total of $9$ bins.

\mysection{Training Details.}
We use the offline training scheme from \cite{shi2019pointrcnn} where RPN proposals and features are pre-computed and saved.
Data augmentation is performed by artificially adding vehicles to the scene along with random global rotations and translations.
During training, $300$ RPN proposals are saved for each frame, but we only consider $64$ proposals for the losses.
Proposals are considered positive if they have an IoU greater than $0.6$, and negative if they have an IoU lower than $0.45$.
Only proposals with an IoU of at least $0.55$ are considered for the regression loss.
The network is trained for $50$ epochs on \KITTItrain only, using Adam optimizer, with a batch size of $256$ proposals (\ie $4$ frames), and a learning rate of $0.0002$.
We run all our training on v100 GPUs with 32GB of memory, and our validation and testing on GTX1080Ti GPUs with 12GB of memory.

\subsection{Main Results}

\Table{MainResults} reports the average precision at an IoU of $0.7$ for 3D and BEV object detection on \KITTItest.
We obtain results that are comparable against other published performances.
PointRGCN is on average $2^{nd}$, after the recently proposed method \STD.
We consistently outperform our baseline \PointRCNN, and achieve state-of-the-art on the easy difficulty on BEV detection task with an impressive $2.13\%$ improvement.
We believe the fine-grained information gathered by the graph representation helps in localizing vehicles in the BEV space, but struggles to improve further the height parameters (\ie $r_y$, $r_h$) with respect to \PointRCNN.

\subsection{Ablation Study}

We compare our method with the state-of-the-art on \KITTIval and provide extensive ablation studies for both our R-GCN and C-GCN modules.
In order to demonstrate the effectiveness of our R-GCN and C-GCN modules, we compare each of our modules against \PointRCNN in \Table{AblationModule}.
%
An improvement in performance is obtained by using R-GCN over RCNN on \KITTIval in hard settings.
By combining both modules we obtain a performance increase in the easy and moderate categories in \KITTIval.
Additionally, our full pipeline generalize better than \PointRCNN as shown from the overall increase in performance on \KITTItest.
This aligns with the better generalization capability of current GCN~\cite{li2019deepgcns_journal}.
%
%

\begin{table}[htbp]
	\centering
	\caption{
	\textbf{Ablation for PointRGCN.}
	We evaluate our modules R-GCN and C-GCN singularly and compared with PointRGCN, on \KITTIval. 
	Best results shown in \textbf{bold}. Time in ms.
	}
	\label{tab:AblationModule}
	\begin{tabular}{l||c|c|c||c}
        Method     &  Easy  &  Mode. &  Hard  & Time \\ \hline\hline
        \PointRCNN &\D88.45 &  77.67 &  76.30 & 100 \\ 
        \hline\hline
        PointRGCN  &  88.37	&\D78.54 &  77.60 & 262 \\ \hline
        R-GCN      &  88.08	&  78.45 &\D77.81 &  239 \\ \hline 
        \cite{shi2019pointrcnn} + C-GCN 
                   &  87.56 &  77.58 &  75.94 &  147 \\ \hline 
        \hline
	\end{tabular}
\end{table}

\subsection{Further Ablation on R-GCN}

We further investigate the contribution of R-GCN in \Table{AblationRGCN}, by ablating it using different depths, layer types, residual skip connections, dilations, and input features.

\begin{table}[htbp]
	\centering
	\caption{
	\textbf{Ablation for R-GCN} on \KITTIval set. 
	We validate here the setup of our R-GCN network.
	Our setup and best results in \textbf{bold}.
	Time in ms.
	}
	\label{tab:AblationRGCN}
	\begin{tabular}{l||c|c|c||c}
\textbf{R-GCN} (3D@0.7)&  Easy  &  Mode. &  Hard  &  Time \\ \hline\hline
        PointRGCN      &  88.37	&  78.54 &	77.60 &  262       \\\hline 
        R-GCN alone    &  88.08	&  78.45 &	77.81 &  239       \\\hline 
\hline  1 layers       &  87.29	&  78.09 &  77.34 &  135       \\\hline 
        3 layers       &  87.96	&\D78.48 &\D77.94 &  188       \\\hline 
      \D5 layers       &\D88.08	&  78.45 &  77.81 &  239       \\\hline 
        10 layers      &  83.33	&  76.56 &  75.47 &  447       \\\hline 
\hline  w/ \EdgeConv   &  87.75	&  78.33 &  77.68 &  282       \\\hline 
        w/o residual   &  82.05	&  73.12 &  73.04 &  237       \\\hline 
        w/o dilation   &  82.84	&  73.34 &  73.31 &  232       \\\hline 
        w/o RPN feat.  &  83.18	&  74.36 &  72.83 &  238       \\\hline 
        \hline
	\end{tabular}
\end{table}

\mysection{Depth.}
We test our R-GCN module using $1$, $3$, $5$ and $10$ layers.
As shown in \Table{AblationRGCN}, the R-GCN module is fairly insensitive across different network depths.
The best results are achieved from $3$ to $5$ layers.

\mysection{Network Design.} 
We show that dilation and residual connections are paramount.
Without residual skip connections, the performance drops $5.36\%$.
Li~\etal~\cite{li2019deepgcns_journal} showed that residual skip connection allows the gradient to flow better between layers, improving the smoothness of the GCN convergence.
Without dilation, the performance drops $5.14\%$ on the moderate validation subset.
Dilation provides the necessary receptive field for our GCN to gather enough feature information.
On the other hand, \MRGCN and \EdgeConv perform fairly similarly, with the biggest difference being in memory and speed requirements. 
We have measured EdgeConv to be approximately $40$ms slower than MRGCN when using a $5$-layer network.
Additionally, the $5$-layer network using MRGCN only consumes a maximum of $1840$MB of memory while EdgeConv consumes a maximumum of $2165$MB of memory.

\mysection{Input data.}
We exhibit a drop of $4.16\%$ on the moderate subset by using only point coordinates as features for the nodes of the graph.
RPN features are an important source of semantic information when processing proposal points since they have been trained for semantic segmentation.

\subsection{Further Ablation on C-GCN}

We further investigate the contribution of C-GCN in \Table{AblationCGCN}, by ablating it using different depths, layer types, residual skip connections, dilation, and input features.

\begin{table}[htbp]
	\centering
	\caption{
	\textbf{Ablation for C-GCN} on \KITTIval set.
	We validate here the setup of our C-GCN network.
	Our setup and best results in \textbf{bold}.
	Time in ms.
	}
	\label{tab:AblationCGCN}
	\begin{tabular}{l||c|c|c||c}
\textbf{C-GCN} (3D@0.7) &  Easy  & Mode. & Hard & Time\\ \hline\hline
        PointRGCN      &  88.37 &  78.54 &  77.60 & 262  \\ 
        C-GCN alone    &  87.56 &  77.58 &  75.94 & 147      \\\hline 
\hline  1 layers       &  86.11 &  77.30 &  75.12 & 145      \\\hline 
      \D3 layers       &\D87.56 &\D77.58 &\D75.94 & 147      \\\hline 
        5 layers       &  86.45 &  76.95 &  75.88 & 151      \\\hline 
        10 layers      &  85.67 &  76.56 &  75.08 & 160      \\\hline 
        20 layers      &  82.69 &  75.45 &  72.96 & 173      \\\hline 
        30 layers      &  83.12 &  74.15 &  71.98 & 192      \\\hline 
        40 layers      &  81.44 &  72.22 &  67.48 & 206      \\\hline 
\hline  w/ \MRGCN      &  85.02	&  76.09 &	73.51 & 152      \\\hline 
        w/o residual   &  82.84	&  73.74 &	73.17 & 150      \\\hline 
        w/ dilation    &  87.03	&  77.37 &	76.06 & 150      \\\hline 
        w/ RPN feat.   &  82.89	&  74.05 &	73.74 & 150      \\\hline 
        \hline 
	\end{tabular}
\end{table}

\mysection{Depth.}
We tested our C-GCN modules with up to $40$ layers.
Since the number of nodes is lower than for the R-GCN, adding layers increases the inference time without providing improved performances.
We have thus chosen to use a total of $3$ layers for the C-GCN network in our pipeline.

\mysection{Network Design.} 
We show on \Table{AblationCGCN} that \MRGCN performs slightly less than \EdgeConv, as expected, since MRGCN is a simplification over EdgeConv.
Like for our R-GCN, residual skip connection are necessary to make the GCN converge smoothly, accounting for $3.84\%$ in the moderate difficulty.
Using dilation slightly improves performances on the hard difficulty, but prefer not using dilation since dilation may overflow neighbors for deeper networks.
Using a global feature derived from the RPN point features does not provide further improvements.


\subsection{Qualitative Results}
We show qualitative results of our detection pipeline in \Figure{Qualitative_Detection}.
It can be observed that in some cases our method is able to detect vehicles which are not labeled in the dataset due to their partial visibility, as in \Figure{Qualitative_Detection}(a).
Additionally, we show an improvement over PointRCNN in cases such as \Figure{Qualitative_Detection}(b), where false positives are avoided due to our C-GCN.
Failure cases can be observed in images \Figure{Qualitative_Detection}(c), where false positives occur due to large occlusion of vehicles in heavy traffic and due to vegetation clutter at large distances.
Overall, we are able to provide tight bounding boxes with high precision.
\section{Conclusion}

Current advances in graph convolutional networks have led to better performances in varied 3D computer vision tasks.
This has motivated us to leverage GCNs for the task of 3D vehicle detection, and demonstrate their effectiveness for vehicle detection.
We introduced two novel GCN-based modules for point feature and context aggregation both within and between proposals.
Making use of both modules, we presented a novel pipeline PointRGCN: a two-stage 3D vehicle detection pipeline that introduces graph representations for proposal boxes refinement.
We showed comparable results with recent works, and report a new state-of-the-art on the easy subset in BEV settings for the KITTI dataset.
Still, much work remains to be done in the field of GCNs, from which this work and others stemming from it will benefit.
%


{\small
	\bibliographystyle{ieee_fullname}
	\bibliography{ms}

\begin{thebibliography}{10}\itemsep=-1pt

\bibitem{Chandra_2019_CVPR}
Rohan Chandra, Uttaran Bhattacharya, Aniket Bera, and Dinesh Manocha.
\newblock Traphic: Trajectory prediction in dense and heterogeneous traffic
  using weighted interactions.
\newblock In {\em The IEEE Conference on Computer Vision and Pattern
  Recognition (CVPR)}, June 2019.

\bibitem{Chang_2019_CVPR}
Ming-Fang Chang, John Lambert, Patsorn Sangkloy, Jagjeet Singh, Slawomir Bak,
  Andrew Hartnett, De Wang, Peter Carr, Simon Lucey, Deva Ramanan, and James
  Hays.
\newblock Argoverse: 3d tracking and forecasting with rich maps.
\newblock In {\em The IEEE Conference on Computer Vision and Pattern
  Recognition (CVPR)}, June 2019.

\bibitem{chen2016monocular}
Xiaozhi Chen, Kaustav Kundu, Ziyu Zhang, Huimin Ma, Sanja Fidler, and Raquel
  Urtasun.
\newblock Monocular 3d object detection for autonomous driving.
\newblock In {\em Proceedings of the IEEE Conference on Computer Vision and
  Pattern Recognition}, pages 2147--2156, 2016.

\bibitem{chen20173d}
Xiaozhi Chen, Kaustav Kundu, Yukun Zhu, Huimin Ma, Sanja Fidler, and Raquel
  Urtasun.
\newblock 3d object proposals using stereo imagery for accurate object class
  detection.
\newblock {\em IEEE transactions on pattern analysis and machine intelligence},
  40(5):1259--1272, 2017.

\bibitem{chen2017multi}
Xiaozhi Chen, Huimin Ma, Ji Wan, Bo Li, and Tian Xia.
\newblock Multi-view 3d object detection network for autonomous driving.
\newblock In {\em Proceedings of the IEEE Conference on Computer Vision and
  Pattern Recognition}, pages 1907--1915, 2017.

\bibitem{Chen2019FastPointRCNN}
Yilun Chen, Shu Liu, Xiaoyong Shen, and Jiaya Jia.
\newblock Fast point r-cnn.
\newblock In {\em The IEEE International Conference on Computer Vision (ICCV)},
  October 2019.

\bibitem{Geiger2012KITTI}
Andreas Geiger, Philip Lenz, and Raquel Urtasun.
\newblock Are we ready for autonomous driving? the kitti vision benchmark
  suite.
\newblock In {\em Conference on Computer Vision and Pattern Recognition
  (CVPR)}, 2012.

\bibitem{Giancola_2019_CVPR}
Silvio Giancola, Jesus Zarzar, and Bernard Ghanem.
\newblock Leveraging shape completion for 3d siamese tracking.
\newblock In {\em The IEEE Conference on Computer Vision and Pattern
  Recognition (CVPR)}, June 2019.

\bibitem{Hu_2019_ICCV}
Hou-Ning Hu, Qi-Zhi Cai, Dequan Wang, Ji Lin, Min Sun, Philipp Krahenbuhl,
  Trevor Darrell, and Fisher Yu.
\newblock Joint monocular 3d vehicle detection and tracking.
\newblock In {\em The IEEE International Conference on Computer Vision (ICCV)},
  October 2019.

\bibitem{ku2018joint}
Jason Ku, Melissa Mozifian, Jungwook Lee, Ali Harakeh, and Steven~L Waslander.
\newblock Joint 3d proposal generation and object detection from view
  aggregation.
\newblock In {\em 2018 IEEE/RSJ International Conference on Intelligent Robots
  and Systems (IROS)}, pages 1--8. IEEE, 2018.

\bibitem{lang2019pointpillars}
Alex~H Lang, Sourabh Vora, Holger Caesar, Lubing Zhou, Jiong Yang, and Oscar
  Beijbom.
\newblock Pointpillars: Fast encoders for object detection from point clouds.
\newblock In {\em Proceedings of the IEEE Conference on Computer Vision and
  Pattern Recognition}, pages 12697--12705, 2019.

\bibitem{lehner2019patch}
Johannes Lehner, Andreas Mitterecker, Thomas Adler, Markus Hofmarcher, Bernhard
  Nessler, and Sepp Hochreiter.
\newblock Patch refinement--localized 3d object detection.
\newblock {\em arXiv preprint arXiv:1910.04093}, 2019.

\bibitem{Li2019deepgcn}
Guohao Li, Matthias Muller, Ali Thabet, and Bernard Ghanem.
\newblock Deepgcns: Can gcns go as deep as cnns?
\newblock In {\em The IEEE International Conference on Computer Vision (ICCV)},
  October 2019.

\bibitem{li2019deepgcns_journal}
Guohao Li, Matthias Müller, Guocheng Qian, Itzel~C. Delgadillo, Abdulellah
  Abualshour, Ali Thabet, and Bernard Ghanem.
\newblock Deepgcns: Making gcns go as deep as cnns, 2019.

\bibitem{Li_2019_CVPR}
Peiliang Li, Xiaozhi Chen, and Shaojie Shen.
\newblock Stereo r-cnn based 3d object detection for autonomous driving.
\newblock In {\em The IEEE Conference on Computer Vision and Pattern
  Recognition (CVPR)}, June 2019.

\bibitem{liang2019multi}
Ming Liang, Bin Yang, Yun Chen, Rui Hu, and Raquel Urtasun.
\newblock Multi-task multi-sensor fusion for 3d object detection.
\newblock In {\em Proceedings of the IEEE Conference on Computer Vision and
  Pattern Recognition}, pages 7345--7353, 2019.

\bibitem{maturana2015voxnet}
Daniel Maturana and Sebastian Scherer.
\newblock Voxnet: A 3d convolutional neural network for real-time object
  recognition.
\newblock In {\em 2015 IEEE/RSJ International Conference on Intelligent Robots
  and Systems (IROS)}, pages 922--928. IEEE, 2015.

\bibitem{mousavian20173d}
Arsalan Mousavian, Dragomir Anguelov, John Flynn, and Jana Kosecka.
\newblock 3d bounding box estimation using deep learning and geometry.
\newblock In {\em Proceedings of the IEEE Conference on Computer Vision and
  Pattern Recognition}, pages 7074--7082, 2017.

\bibitem{Nguyen_2019_ICCV}
Anh-Duc Nguyen, Seonghwa Choi, Woojae Kim, and Sanghoon Lee.
\newblock Graphx-convolution for point cloud deformation in 2d-to-3d
  conversion.
\newblock In {\em The IEEE International Conference on Computer Vision (ICCV)},
  October 2019.

\bibitem{Qi_2019_ICCV}
Charles~R. Qi, Or Litany, Kaiming He, and Leonidas~J. Guibas.
\newblock Deep hough voting for 3d object detection in point clouds.
\newblock In {\em The IEEE International Conference on Computer Vision (ICCV)},
  October 2019.

\bibitem{qi2018frustum}
Charles~R Qi, Wei Liu, Chenxia Wu, Hao Su, and Leonidas~J Guibas.
\newblock Frustum pointnets for 3d object detection from rgb-d data.
\newblock In {\em Proceedings of the IEEE Conference on Computer Vision and
  Pattern Recognition}, pages 918--927, 2018.

\bibitem{qi2017pointnet}
Charles~R Qi, Hao Su, Kaichun Mo, and Leonidas~J Guibas.
\newblock Pointnet: Deep learning on point sets for 3d classification and
  segmentation.
\newblock In {\em Proceedings of the IEEE Conference on Computer Vision and
  Pattern Recognition}, pages 652--660, 2017.

\bibitem{qi2016volumetric}
Charles~R Qi, Hao Su, Matthias Nie{\ss}ner, Angela Dai, Mengyuan Yan, and
  Leonidas~J Guibas.
\newblock Volumetric and multi-view cnns for object classification on 3d data.
\newblock In {\em Proceedings of the IEEE conference on computer vision and
  pattern recognition}, pages 5648--5656, 2016.

\bibitem{qi2017pointnet++}
Charles~Ruizhongtai Qi, Li Yi, Hao Su, and Leonidas~J Guibas.
\newblock Pointnet++: Deep hierarchical feature learning on point sets in a
  metric space.
\newblock In {\em Advances in neural information processing systems}, pages
  5099--5108, 2017.

\bibitem{riegler2017octnet}
Gernot Riegler, Ali Osman~Ulusoy, and Andreas Geiger.
\newblock Octnet: Learning deep 3d representations at high resolutions.
\newblock In {\em Proceedings of the IEEE Conference on Computer Vision and
  Pattern Recognition}, pages 3577--3586, 2017.

\bibitem{shi2019pointrcnn}
Shaoshuai Shi, Xiaogang Wang, and Hongsheng Li.
\newblock Pointrcnn: 3d object proposal generation and detection from point
  cloud.
\newblock In {\em Proceedings of the IEEE Conference on Computer Vision and
  Pattern Recognition}, pages 770--779, 2019.

\bibitem{shi2019part}
Shaoshuai Shi, Zhe Wang, Xiaogang Wang, and Hongsheng Li.
\newblock Part-a\^{} 2 net: 3d part-aware and aggregation neural network for
  object detection from point cloud.
\newblock {\em arXiv preprint arXiv:1907.03670}, 2019.

\bibitem{Shu_2019_ICCV}
Dong~Wook Shu, Sung~Woo Park, and Junseok Kwon.
\newblock 3d point cloud generative adversarial network based on tree
  structured graph convolutions.
\newblock In {\em The IEEE International Conference on Computer Vision (ICCV)},
  October 2019.

\bibitem{su2015multi}
Hang Su, Subhransu Maji, Evangelos Kalogerakis, and Erik Learned-Miller.
\newblock Multi-view convolutional neural networks for 3d shape recognition.
\newblock In {\em Proceedings of the IEEE international conference on computer
  vision}, pages 945--953, 2015.

\bibitem{tatarchenko2018tangent}
Maxim Tatarchenko, Jaesik Park, Vladlen Koltun, and Qian-Yi Zhou.
\newblock Tangent convolutions for dense prediction in 3d.
\newblock In {\em Proceedings of the IEEE Conference on Computer Vision and
  Pattern Recognition}, pages 3887--3896, 2018.

\bibitem{wang2019voxel}
Bei Wang, Jianping An, and Jiayan Cao.
\newblock Voxel-fpn: multi-scale voxel feature aggregation in 3d object
  detection from point clouds.
\newblock {\em arXiv preprint arXiv:1907.05286}, 2019.

\bibitem{Wang_2019_CVPR}
Lei Wang, Yuchun Huang, Yaolin Hou, Shenman Zhang, and Jie Shan.
\newblock Graph attention convolution for point cloud semantic segmentation.
\newblock In {\em The IEEE Conference on Computer Vision and Pattern
  Recognition (CVPR)}, June 2019.

\bibitem{wang2019pseudo}
Yan Wang, Wei-Lun Chao, Divyansh Garg, Bharath Hariharan, Mark Campbell, and
  Kilian~Q Weinberger.
\newblock Pseudo-lidar from visual depth estimation: Bridging the gap in 3d
  object detection for autonomous driving.
\newblock In {\em Proceedings of the IEEE Conference on Computer Vision and
  Pattern Recognition}, pages 8445--8453, 2019.

\bibitem{wang2019deep}
Yue Wang and Justin~M Solomon.
\newblock Deep closest point: Learning representations for point cloud
  registration.
\newblock {\em arXiv preprint arXiv:1905.03304}, 2019.

\bibitem{wang2019dynamic}
Yue Wang, Yongbin Sun, Ziwei Liu, Sanjay~E Sarma, Michael~M Bronstein, and
  Justin~M Solomon.
\newblock Dynamic graph cnn for learning on point clouds.
\newblock {\em ACM Transactions on Graphics (TOG)}, 38(5):146, 2019.

\bibitem{wu20153d}
Zhirong Wu, Shuran Song, Aditya Khosla, Fisher Yu, Linguang Zhang, Xiaoou Tang,
  and Jianxiong Xiao.
\newblock 3d shapenets: A deep representation for volumetric shapes.
\newblock In {\em Proceedings of the IEEE conference on computer vision and
  pattern recognition}, pages 1912--1920, 2015.

\bibitem{xu2018multi}
Bin Xu and Zhenzhong Chen.
\newblock Multi-level fusion based 3d object detection from monocular images.
\newblock In {\em Proceedings of the IEEE Conference on Computer Vision and
  Pattern Recognition}, pages 2345--2353, 2018.

\bibitem{yan2018second}
Yan Yan, Yuxing Mao, and Bo Li.
\newblock Second: Sparsely embedded convolutional detection.
\newblock {\em Sensors}, 18(10):3337, 2018.

\bibitem{yang2018pixor}
Bin Yang, Wenjie Luo, and Raquel Urtasun.
\newblock Pixor: Real-time 3d object detection from point clouds.
\newblock In {\em Proceedings of the IEEE conference on Computer Vision and
  Pattern Recognition}, pages 7652--7660, 2018.

\bibitem{Yang2019std}
Zetong Yang, Yanan Sun, Shu Liu, Xiaoyong Shen, and Jiaya Jia.
\newblock Std: Sparse-to-dense 3d object detector for point cloud.
\newblock In {\em The IEEE International Conference on Computer Vision (ICCV)},
  October 2019.

\bibitem{yu2015multi}
Fisher Yu and Vladlen Koltun.
\newblock Multi-scale context aggregation by dilated convolutions.
\newblock {\em arXiv preprint arXiv:1511.07122}, 2015.

\bibitem{zhou2018voxelnet}
Yin Zhou and Oncel Tuzel.
\newblock Voxelnet: End-to-end learning for point cloud based 3d object
  detection.
\newblock In {\em Proceedings of the IEEE Conference on Computer Vision and
  Pattern Recognition}, pages 4490--4499, 2018.

\end{thebibliography}
}

\begin{figure*}[t]
    \centering
    \begin{tabular}{c|c|c}
    \Large \PointRCNN & \Large R-GCN alone (ours) & \Large PointRGCN (ours) \\ \midrule
    \begin{overpic}[width=0.30\linewidth,trim={1.8cm 0.9cm 1.4cm 1.2cm},clip]
    {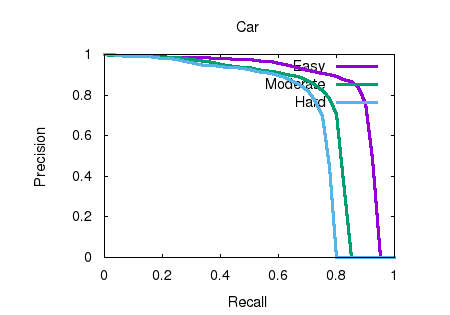}
        \put (20,43){\textbf{3D@0.7}}
        \put (20,36){\textbf{PointRCNN}}
        \put (20,29){\color{Purple}   \textbf{Easy: 85.94}}
        \put (20,22){\color{Green}    \textbf{Mode:75.76}} 
        \put (20,15){\color{RoyalBlue}\textbf{Hard: 68.32}}
    \end{overpic} &
    \begin{overpic}[width=0.30\linewidth,trim={1.8cm 0.9cm 1.4cm 1.2cm},clip]
    {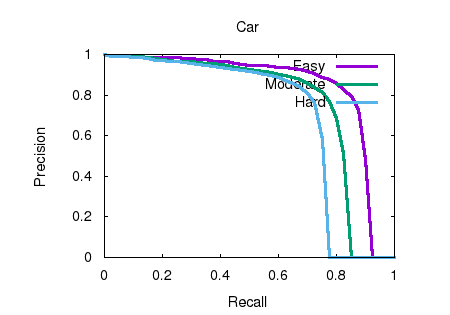}
        \put (20,43){\textbf{3D@0.7}}
        \put (20,36){\textbf{R-GCN}}
        \put (20,29){\color{Purple}   \textbf{Easy: 83.42}}
        \put (20,22){\color{Green}    \textbf{Mode:75.26}}
        \put (20,15){\color{RoyalBlue}\textbf{Hard: 68.73}}
    \end{overpic} &
    \begin{overpic}[width=0.30\linewidth,trim={1.8cm 0.9cm 1.4cm 1.2cm},clip]
    {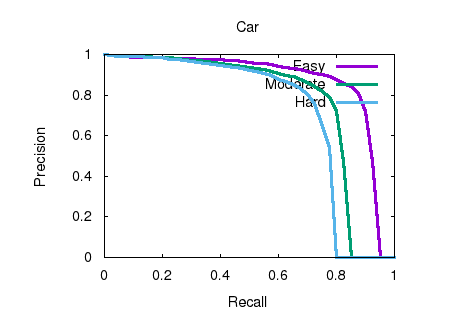}
        \put (20,43){\textbf{3D@0.7}}
        \put (20,36){\textbf{PointRGCN}}
        \put (20,29){\color{Purple}   \textbf{Easy: 85.97}}
        \put (20,22){\color{Green}    \textbf{Mode:75.73}}
        \put (20,15){\color{RoyalBlue}\textbf{Hard: 70.60}}
    \end{overpic} \\ \midrule


    \begin{overpic}[width=0.30\linewidth,trim={1.8cm 0.9cm 1.4cm 1.2cm},clip]
    {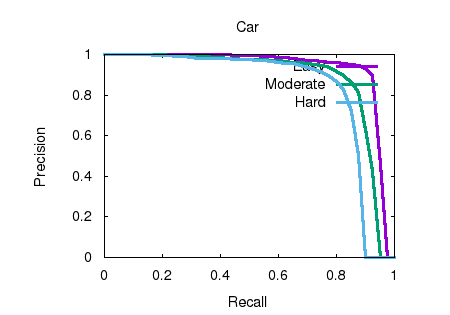}
        \put (20,43){\textbf{BEV@0.7}}
        \put (20,36){\textbf{PointRCNN}}
        \put (20,29){\color{Purple}   \textbf{Easy: 89.47}}
        \put (20,22){\color{Green}    \textbf{Mode:85.68}} 
        \put (20,15){\color{RoyalBlue}\textbf{Hard: 79.10}}
    \end{overpic} &
    \begin{overpic}[width=0.30\linewidth,trim={1.8cm 0.9cm 1.4cm 1.2cm},clip]
    {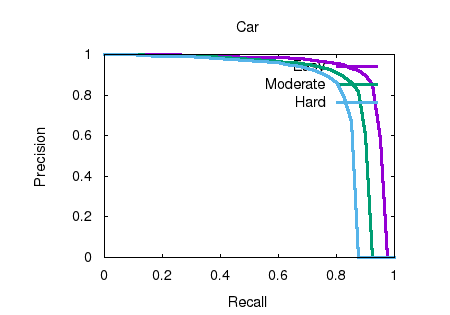}
        \put (20,43){\textbf{BEV@0.7}}
        \put (20,36){\textbf{R-GCN}}
        \put (20,29){\color{Purple}   \textbf{Easy: 91.91}}
        \put (20,22){\color{Green}    \textbf{Mode:86.05}}
        \put (20,15){\color{RoyalBlue}\textbf{Hard: 81.05}}
    \end{overpic} &
    \begin{overpic}[width=0.30\linewidth,trim={1.8cm 0.9cm 1.4cm 1.2cm},clip]
    {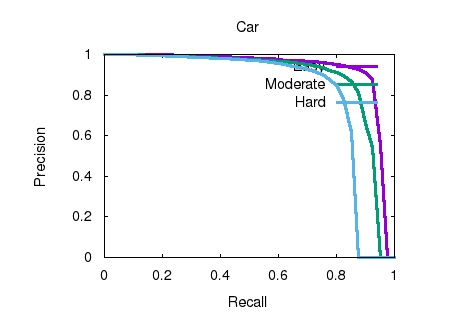}
        \put (20,43){\textbf{BEV@0.7}}
        \put (20,36){\textbf{PointRGCN}}
        \put (20,29){\color{Purple}   \textbf{Easy: 91.63}}
        \put (20,22){\color{Green}    \textbf{Mode:87.49}}
        \put (20,15){\color{RoyalBlue}\textbf{Hard: 80.73}}
    \end{overpic} \\ \midrule

    \begin{overpic}[width=0.30\linewidth,trim={1.8cm 0.9cm 1.4cm 1.2cm},clip]
    {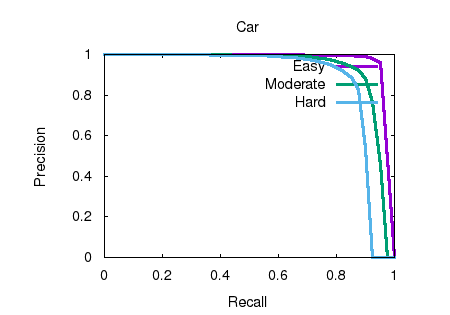}
        \put (20,43){\textbf{2D@0.7}}
        \put (20,36){\textbf{PointRCNN}}
        \put (20,29){\color{Purple}   \textbf{Easy: 95.92}}
        \put (20,22){\color{Green}    \textbf{Mode:91.90}} 
        \put (20,15){\color{RoyalBlue}\textbf{Hard: 87.11}}
    \end{overpic} &
    \begin{overpic}[width=0.30\linewidth,trim={1.8cm 0.9cm 1.4cm 1.2cm},clip]
    {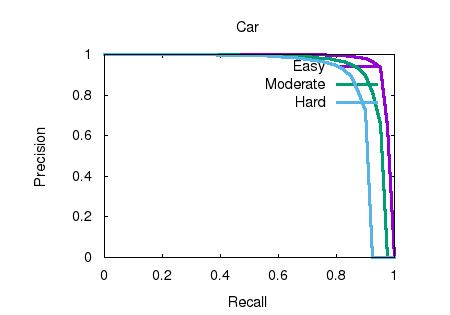}
        \put (20,43){\textbf{2D@0.7}}
        \put (20,36){\textbf{R-GCN}}
        \put (20,29){\color{Purple}   \textbf{Easy: 96.19}} 
        \put (20,22){\color{Green}    \textbf{Mode:92.67}}
        \put (20,15){\color{RoyalBlue}\textbf{Hard: 87.66}}
    \end{overpic} &
    \begin{overpic}[width=0.30\linewidth,trim={1.8cm 0.9cm 1.4cm 1.2cm},clip]
    {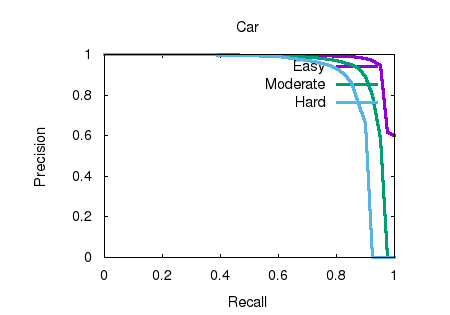}
        \put (20,43){\textbf{2D@0.7}}
        \put (20,36){\textbf{PointRGCN}}
        \put (20,29){\color{Purple}   \textbf{Easy: 97.51}}
        \put (20,22){\color{Green}    \textbf{Mode:92.33}}
        \put (20,15){\color{RoyalBlue}\textbf{Hard: 87.07}}
    \end{overpic} \\ \midrule
    
    \begin{overpic}[width=0.30\linewidth,trim={1.8cm 0.9cm 1.4cm 1.2cm},clip]
    {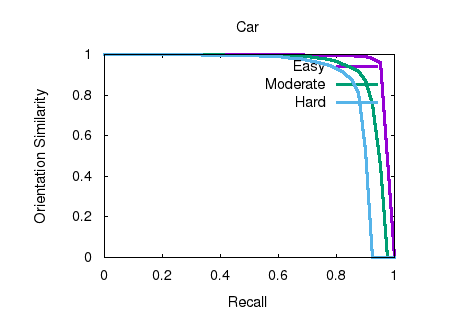}
        \put (20,43){\textbf{AOS}}
        \put (20,36){\textbf{PointRCNN}}
        \put (20,29){\color{Purple}   \textbf{Easy: 95.90}}
        \put (20,22){\color{Green}    \textbf{Mode:91.77}}
        \put (20,15){\color{RoyalBlue}\textbf{Hard: 86.92}}
    \end{overpic} &
    \begin{overpic}[width=0.30\linewidth,trim={1.8cm 0.9cm 1.4cm 1.2cm},clip]
    {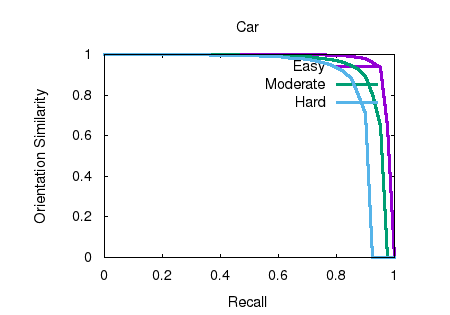}
        \put (20,43){\textbf{AOS}}
        \put (20,36){\textbf{R-GCN}}
        \put (20,29){\color{Purple}   \textbf{Easy: 96.16}} 
        \put (20,22){\color{Green}    \textbf{Mode:92.53}}
        \put (20,15){\color{RoyalBlue}\textbf{Hard: 87.45}}
    \end{overpic} &
    \begin{overpic}[width=0.30\linewidth,trim={1.8cm 0.9cm 1.4cm 1.2cm},clip]
    {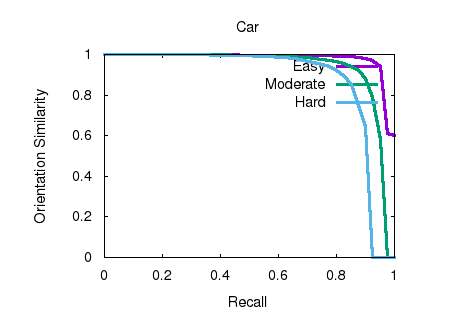}
        \put (20,43){\textbf{AOS}}
        \put (20,36){\textbf{PointRGCN}}
        \put (20,29){\color{Purple}   \textbf{Easy: 97.48}} 
        \put (20,22){\color{Green}    \textbf{Mode:92.15}}
        \put (20,15){\color{RoyalBlue}\textbf{Hard: 86.83}}
    \end{overpic}
	\end{tabular}
    \caption{
    \textbf{Precision Recall Curves} the official \KITTItest leaderboard for \PointRCNN, our R-GCN alone and our PointRGCN.
    \textbf{Top to bottom:} 3D@0.7, BEV@0.7, 2D@0.7 and AOS. 
    For \PointRCNN, 3D@0.7 and BEV@0.7 are taken from their paper, but since they do not report 2D@0.7 and AOS we took the best entry on the KITTI leaderboard.
    }
	\label{fig:KITTIleaderboard}
\end{figure*}
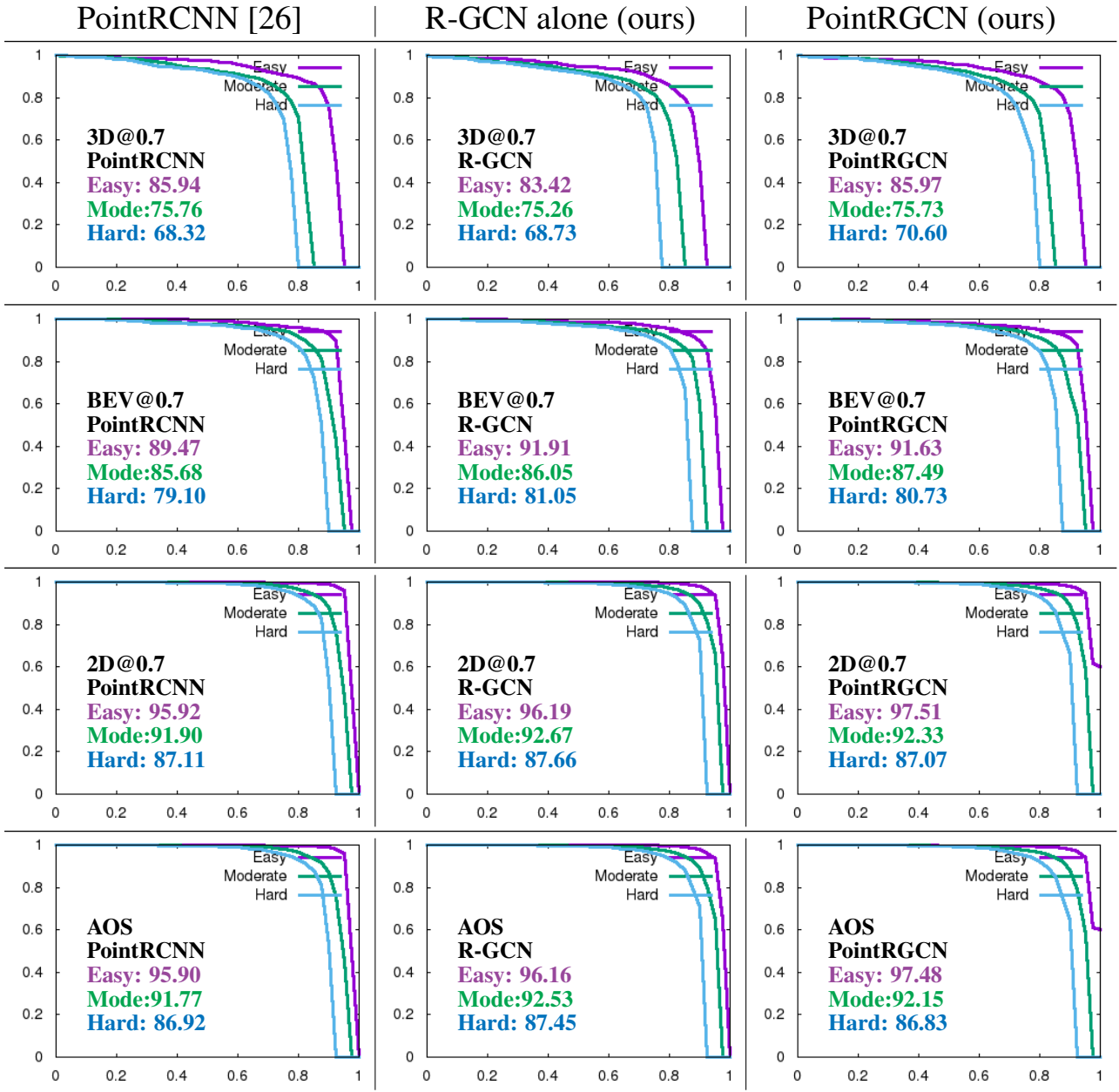



\newpage

\section{Supplementary Material.}
We present in \Figure{KITTIleaderboard} the Precision-Recall curves from the KITTI leaderboard, established on the \emph{testing} set.
In particular, we compare the results of PointRCNN, our R-GCN module alone, and our PointRGCN pipeline (from left to right).
We refer to the 3D@0.7, BEV@0.7, 2D@0.7 and AOS metrics (from top to bottom).
Note how our model improves the 2D@0.7 and AOS metrics on the Easy difficulty setting, reaching a recall of 100\%.
This shows that we are able to find \textbf{\emph{all easy vehicle samples}} in the 2D RGB frame with a precision of 60\%. 
Such performances were previously unseen with PointRCNN.

To enable further comparison, we provide the ablation study of our method on BEV, 2D Bounding Boxes and AOS.
We complement the ablation study on R-GCN from \Table{AblationRGCN} in \Table{AblationRGCN_complement}
and on C-GCN from \Table{AblationCGCN} in \Table{AblationCGCN_complement}.


\begin{table*}[t]
	\centering
	\caption{
	\textbf{Ablation for R-GCN} on \KITTIval set. 
	We validate here the setup of our R-GCN network.
	Our setup and best results in \textbf{bold}.
	Time in ms.
	}
	\label{tab:AblationRGCN_complement}
	\begin{tabular}{l||c|c|c||c|c|c||c|c|c||c}
         &  \multicolumn{3}{c||}{BEV @ 0.7 IoU}&  \multicolumn{3}{c||}{2D @ 0.7 IoU}&  \multicolumn{3}{c||}{AOS} & Time \\ \hline
\textbf{R-GCN}          &  Easy  &  Mode. &  Hard  &  Easy  &  Mode. &  Hard  &  Easy  &  Mode. &  Hard  &  (ms) \\ \hline
\hline  PointRGCN       &  89.68 &  87.71 &  85.66 &  96.19	&  89.71 &  88.58 &  96.18 &  89.58 &  88.35 &  262       \\\hline 
        R-GCN alone     &  89.76 &  87.80 &  86.11 &  96.22	&  89.59 &  88.94 &  96.21 &  89.45 &  88.73 &  239       \\\hline 
\hline  1 layers        &  89.40 &  87.59 &  85.86 &  90.36	&  89.38 &  88.65 &  90.35 &  89.23 &  88.43 &  135       \\\hline 
        3 layers        &  89.67 &  87.77 &\D86.25 &\D96.67	&  89.46 &  88.84 &\D96.66 &  89.32 &  88.63 &  188       \\\hline 
      \D5 layers        &\D89.76 &\D87.80 &  86.11 &  96.22	&\D89.59 &\D88.94 &  96.21 &\D89.45 &\D88.73 &  239       \\\hline 
        10 layers       &  89.41 &  87.57 &  86.07 &  90.31	&  89.19 &  88.45 &  90.30 &  89.05 &  88.23 &  447       \\\hline 
\hline  w/ \EdgeConv    &  89.41 &  87.69 &  85.97 &  96.29	&  89.61 &  88.91 &  96.28 &  89.44 &  88.67 &  282       \\\hline 
        w/o residual    &  87.04 &  84.53 &  78.03 &  89.09	&  87.52 &  87.11 &  89.08 &  87.33 &  86.82 &  237       \\\hline 
        w/o dilation    &  87.84 &  84.70 &  78.31 &  89.60	&  87.56 &  87.11 &  89.59 &  87.36 &  86.81 &  232       \\\hline 
        w/o RPN feat.   &  89.41 &  87.39 &  85.51 &  90.31	&  88.96 &  88.07 &  90.30 &  88.77 &  87.83 &  238       \\\hline 
        \hline
	\end{tabular}
\end{table*}

\begin{table*}[t]
	\centering
	\caption{
	\textbf{Ablation for C-GCN} on \KITTIval set.
	We validate here the setup of our C-GCN network.
	Our setup and best results in \textbf{bold}.
	Time in ms.
	}
	\label{tab:AblationCGCN_complement}
	\begin{tabular}{l||c|c|c||c|c|c||c|c|c||c}
         &  \multicolumn{3}{c||}{BEV @ 0.7 IoU}&  \multicolumn{3}{c||}{2D @ 0.7 IoU}&  \multicolumn{3}{c||}{AOS} & Time \\ \hline
\textbf{C-GCN}           &  Easy  &  Mode. &  Hard  &  Easy  &  Mode. &  Hard  &  Easy  &  Mode. &  Hard  & (ms)\\ \hline\hline
        PointRGCN        &  89.68 &	 87.71 &  85.66 &  96.19 &  89.71 &  88.58 &  96.18 &  89.58 &  88.35 & 262  \\ \hline
        C-GCN alone      &  89.57 &	 86.86 &  84.90 &  96.16 &  89.36 &  87.67 &  96.15 &  89.21 &  87.42 & 147      \\\hline 
\hline  1 layers         &\D89.61 &\D87.35 &\D85.43 &  90.66 &\D89.48 &\D88.43 &  90.65 &\D89.35 &\D88.19 & 145      \\\hline 
      \D3 layers         &  89.57 &	 86.86 &  84.90 &\D96.16 &  89.36 &  87.67 &\D96.15 &  89.21 &  87.42 & 147      \\\hline 
        5 layers         &  89.35 &	 86.69 &  84.90 &  95.17 &  89.23 &  87.60 &  95.16 &  89.05 &  87.33 & 151      \\\hline 
        10 layers        &  89.30 &	 86.33 &  85.21 &  90.57 &  89.25 &  87.38 &  90.56 &  89.12 &  87.16 & 160      \\\hline 
        20 layers        &  88.95 &	 85.80 &  84.75 &  90.64 &  89.17 &  86.97 &  90.63 &  89.02 &  86.72 & 173      \\\hline 
        30 layers        &  87.57 &	 85.11 &  83.95 &  89.90 &  88.29 &  86.21 &  89.89 &  88.11 &  85.91 & 192      \\\hline 
        40 layers        &  89.47 &	 86.47 &  85.15 &  90.41 &  88.64 &  86.15 &  90.41 &  88.48 &  85.91 & 206      \\\hline 
\hline  w/ \MRGCN        &  89.19 &	 86.83 &  85.21 &  90.59 &  89.30 &  87.35 &  90.59 &  89.13 &  87.08 & 152      \\\hline 
        w/o residual     &  87.29 &	 84.46 &  78.22 &  89.39 &  87.70 &  87.03 &  89.38 &  87.47 &  86.71 & 150      \\\hline 
        w/ dilation      &  89.32 &	 86.92 &  85.21 &  95.72 &  89.47 &  87.97 &  95.71 &  89.32 &  87.73 & 150      \\\hline 
        w/ RPN feat.     &  87.81 &	 84.96 &  78.73 &  89.75 &  87.96 &  87.50 &  89.75 &  87.72 &  87.15 & 150      \\\hline 
        \hline 
	\end{tabular}
\end{table*}


\end{document}